\documentclass[10pt,journal,compsoc]{IEEEtran}
\ifCLASSOPTIONcompsoc
  \usepackage[nocompress]{cite}
\else
  \usepackage{cite}
\fi
\ifCLASSINFOpdf
\else
\fi
\usepackage{color,xcolor}
\usepackage{epsfig}
\usepackage{graphicx}

\usepackage[export]{adjustbox}
\usepackage{array}
\usepackage{booktabs}
\usepackage{colortbl}
\usepackage{float,wrapfig}
\usepackage{hhline}
\usepackage{multirow}
\usepackage{ragged2e}

\usepackage{amsmath,amsfonts,amsthm,amssymb}
\usepackage{bm}
\usepackage{nicefrac}
\usepackage{microtype}

\usepackage{changepage}
\usepackage{extramarks}
\usepackage{fancyhdr}
\usepackage{lastpage}
\usepackage{setspace}
\usepackage{subcaption}
\usepackage{soul}
\usepackage{xspace}
\usepackage{amsfonts}
\usepackage{adjustbox}
\usepackage{multirow}
\usepackage{pifont}
\usepackage{enumitem}

\usepackage{url}

\usepackage{algorithm, algorithmic}
\usepackage{enumerate}
\usepackage{todonotes}

\usepackage{bbm}

\newcommand{\cmark}{\checkmark}

\newcolumntype{L}[1]{>{\raggedright\let\newline\\\arraybackslash\hspace{0pt}}m{#1}}
\newcolumntype{C}[1]{>{\centering\let\newline\\\arraybackslash\hspace{0pt}}m{#1}}
\newcolumntype{R}[1]{>{\raggedleft\let\newline\\\arraybackslash\hspace{0pt}}m{#1}}

\newcommand{\ignorethis}[1]{}

\makeatletter
\DeclareRobustCommand\onedot{\futurelet\@let@token\@onedot}
\def\@onedot{\ifx\@let@token.\else.\null\fi\xspace}

\def\eg{\emph{e.g}\onedot} 
\def\ie{\emph{i.e}\onedot} 
 
\def\etc{\emph{etc}\onedot} 
 
 \def\etal{\emph{et al}\onedot}
\makeatother

\definecolor{MyDarkBlue}{rgb}{0,0.08,1}
\definecolor{MyDarkGreen}{rgb}{0.02,0.6,0.02}
\definecolor{MyDarkRed}{rgb}{0.8,0.02,0.02}
\definecolor{MyDarkOrange}{rgb}{0.40,0.2,0.02}
\definecolor{MyPurple}{RGB}{111,0,255}
\definecolor{MyRed}{rgb}{1.0,0.0,0.0}
\definecolor{MyGold}{rgb}{0.75,0.6,0.12}
\definecolor{MyDarkgray}{rgb}{0.66, 0.66, 0.66}

\newcommand{\myparagraph}[1]{\vspace{5pt}\noindent\textbf{#1}\hspace{10pt}}

\newcommand{\PQ}{PQ}
\newcommand{\PQda}{PQ\textsuperscript{$\dagger$}}
\newcommand{\RQ}{RQ}
\newcommand{\SQ}{SQ}
\newcommand{\PQth}{PQ\textsuperscript{Th}}
\newcommand{\RQth}{RQ\textsuperscript{Th}}
\newcommand{\SQth}{SQ\textsuperscript{Th}}
\newcommand{\PQst}{PQ\textsuperscript{St}}
\newcommand{\RQst}{RQ\textsuperscript{St}}
\newcommand{\SQst}{SQ\textsuperscript{St}}
\newcommand{\miou}{mIoU}

\begin{document}
%
\title{Cylindrical and Asymmetrical 3D Convolution Networks for LiDAR-based Perception}
%
%
%
%


\author{Xinge~Zhu, Hui~Zhou, Tai~Wang, Fangzhou~Hong, Wei~Li,\\ Yuexin~Ma, Hongsheng~Li, Ruigang~Yang, Dahua~Lin

\IEEEcompsocitemizethanks{
\IEEEcompsocthanksitem Xinge Zhu, Tai Wang, Hongsheng Li and Dahua Lin are with the Chinese University of Hong Kong. E-mail: \{zx018, wt019, dhlin\}@ie.cuhk.edu.hk, hsli@ee.cuhk.edu.hk.
\IEEEcompsocthanksitem Hui Zhou is with SenseTime Research. E-mail: smarthuizhou@gmail.com.
\IEEEcompsocthanksitem Yuexin Ma is with ShanghaiTech University, Shanghai Engineering Research Center of Intelligent Vision and Imaging. E-mail: mayuexin@shanghaitech.edu.cn.
\IEEEcompsocthanksitem Wei Li is with Peking University. E-mail: liweimcc@gmail.com
\IEEEcompsocthanksitem Fanzhou Hong is with Nanyang Technological University. E-mail: fangzhouhong820@gmail.com
\IEEEcompsocthanksitem Ruigang Yang is with University of Kentucky. E-mail: ryang@cs.uky.edu
}
}

\markboth{IEEE TRANSACTIONS ON PATTERN ANALYSIS AND MACHINE INTELLIGENCE, VOL. X, NO. X, MMMMMMM YYYY}%
{ZHU \MakeLowercase{\textit{et al.}}: Cylindrical and Asymmetrical 3D Convolution Networks for LiDAR-based Perception}

%



\IEEEtitleabstractindextext{%
\begin{abstract}
    
State-of-the-art methods for driving-scene LiDAR-based perception (including point cloud semantic segmentation, panoptic segmentation and 3D detection, \etc) often project the point clouds to 2D space and then process them via 2D convolution. 
Although this cooperation shows the competitiveness in the point cloud, it inevitably alters and abandons the 3D topology and geometric relations. A natural remedy is to utilize the 3D voxelization and 3D convolution network.
However, we found that in the outdoor point cloud, the improvement obtained in this way is quite limited. An important reason is the property of the outdoor point cloud, namely sparsity and varying density.
Motivated by this investigation, we propose a new framework for the outdoor LiDAR segmentation, where cylindrical partition and asymmetrical 3D convolution networks are designed to explore the 3D geometric pattern while maintaining these inherent properties. 
The proposed model acts as a backbone and the learned features from this model can be used for downstream tasks such as point cloud semantic and panoptic segmentation or 3D detection. In this paper, we benchmark our model on these three tasks.
For semantic segmentation, we evaluate the proposed model on several large-scale datasets, \ie, SemanticKITTI, nuScenes and A2D2. Our method achieves the state-of-the-art on the leaderboard of SemanticKITTI (both single-scan and multi-scan challenge), and significantly outperforms existing methods on nuScenes and A2D2 dataset. Furthermore, the proposed 3D framework also shows strong performance and good generalization on LiDAR panoptic segmentation and LiDAR 3D detection.

\end{abstract}

\begin{IEEEkeywords}
cylindrical partition,  asymmetrical convolution,  point cloud semantic segmentation, point cloud 3D detection, point cloud panoptic segmentation
\end{IEEEkeywords}}

\maketitle

\IEEEdisplaynontitleabstractindextext

%
\IEEEpeerreviewmaketitle

\IEEEraisesectionheading{\section{Introduction}\label{sec:introduction}}	 \IEEEPARstart{3}D {LiDAR} sensor has become an {indispensable device} in {modern} autonomous driving {vehicles}~\cite{ma2019trafficpredict}. It {captures} more precise and farther-away {distance measurements~\cite{xu2019depth} of the surrounding environments} than conventional visual cameras~\cite{wang2021probabilistic,wang2021fcos3d,peng2021side}. The measurements of the sensor naturally form 3D point clouds that can be used to realize a thorough scene understanding for autonomous driving planning and execution, in which LiDAR-based segmentation and detection are crucial for driving-scene perception and understanding.
	
    
    \begin{figure*}
    \centering
    \begin{subfigure}[t]{0.6\textwidth}
    \centering
\includegraphics[width=1\linewidth]{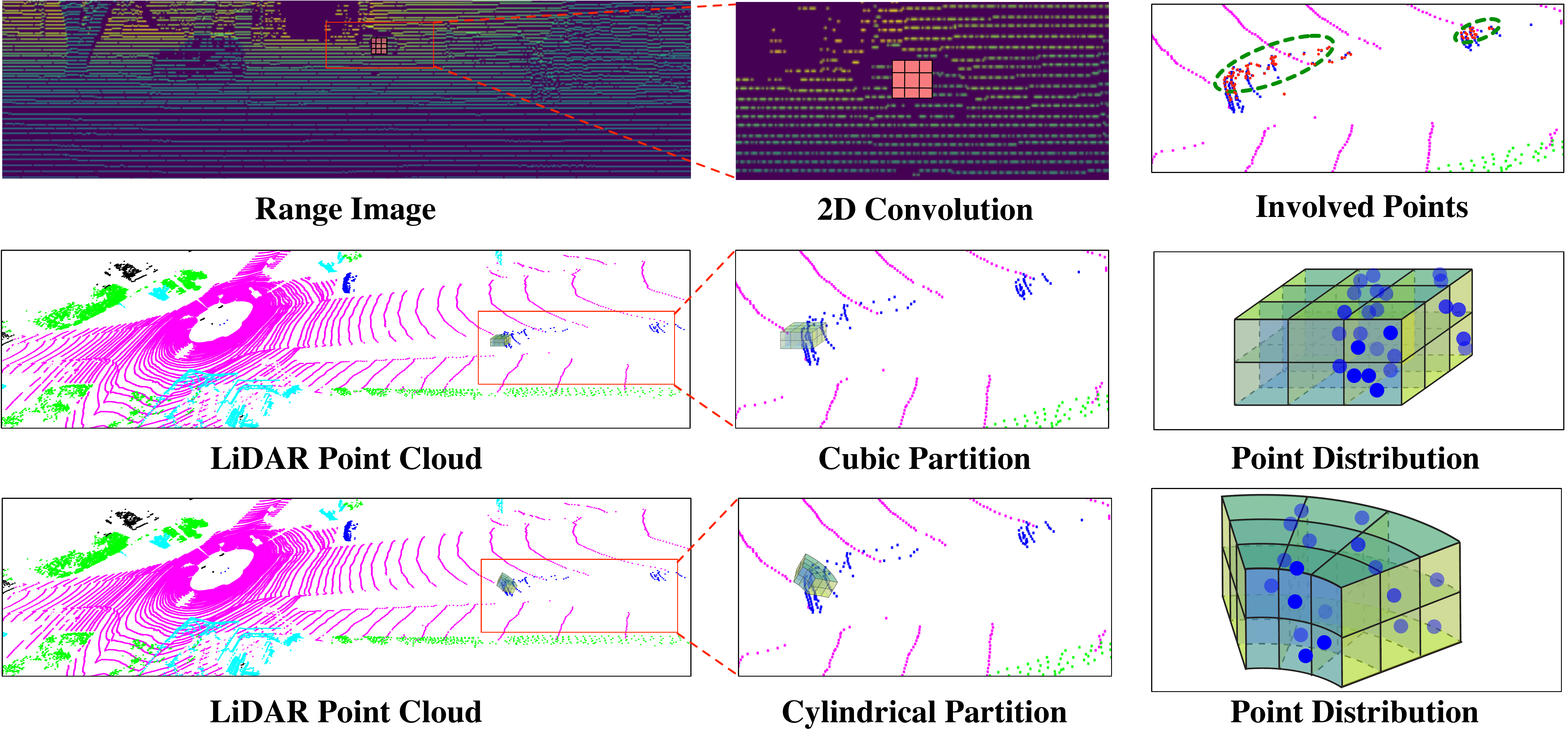}
    \caption{}
    \label{fig:2d_3d_conv}
    \end{subfigure}
    \hfil
    \begin{subfigure}[t]{0.3\textwidth}        \includegraphics[width=\linewidth]{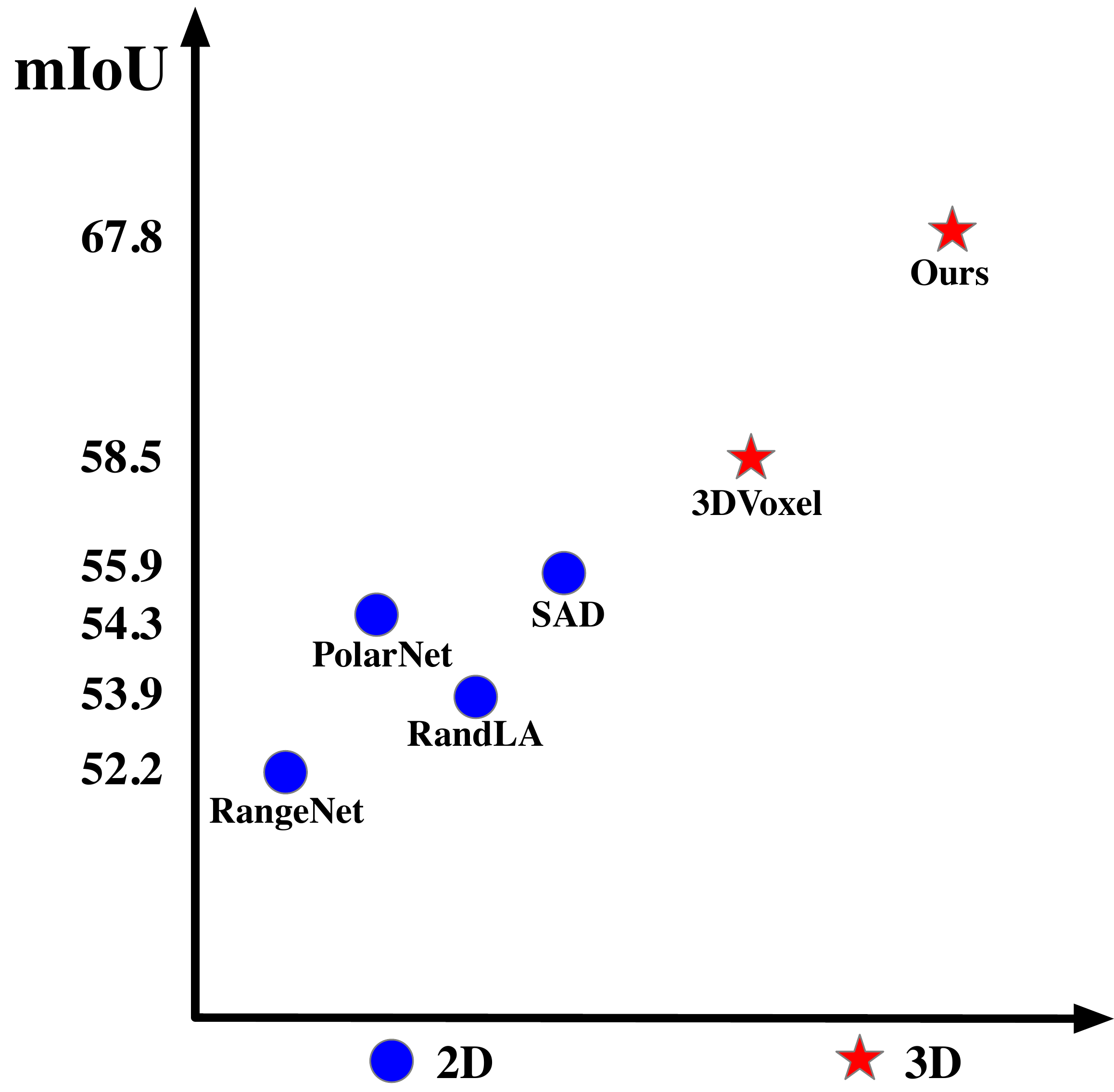}
    \caption{}
    \label{fig:stat1}
    \end{subfigure}
    \vspace{-1ex}
    \caption{(a) Range Image (2D projection) v.s. Cubic Partition v.s. Cylindrical Partition. From top row, it can be found that range image abandons the 3D topology, where 2d convolution processes points in different locations (far away from each other in green circles). From bottom part, cylindrical partition generates the more balanced point distribution than cubic partition (89\% v.s. 61\% cells containing points). (b) Applying the regular 3D voxel partition and 3D convolution directly (\ie, 3DVoxel) gets limited performance gain compared to projection-based (2D) methods~\cite{zhang2020polarnet, hu2020randla, milioto2019rangenet++, xu2020squeezesegv3}, while our method achieves a remarkable performance gain by further tackling the inherent difficulty of outdoor LiDAR point clouds (showing results on SemanticKITTI dataset).
    }
    \label{fig:teaser}
    \vspace{-3ex}
    \end{figure*}
    
    Recently, the advances in deep learning have significantly pushed forward the state of the art in image domain such as image segmentation and detection. Some existing LiDAR-based perception approaches follow this route to project the 3D point clouds onto a 2D space and process them via 2D convolution networks, including range image based~\cite{milioto2019rangenet++,wu2018squeezeseg} and bird's-eye-view
   image based~\cite{zhang2020polarnet,lang2019pointpillars}. However, this group of methods lose and alter the accurate 3D geometric information during the 3D-to-2D projection (as shown in the top row of Fig.~\ref{fig:2d_3d_conv}).
   
   A natural alternative is to utilize the 3D partition and 3D convolution networks to process the point cloud and maintain their 3D geometric relations. However, in our initial attempts, we directly apply the 3D voxelization~\cite{graham20183d, cciccek20163d} and 3D convolution networks to outdoor LiDAR point cloud, only to find very limited performance gain (as shown in Fig.~\ref{fig:stat1}).
   Our investigation into this issue reveals a key difficulty of outdoor LiDAR point cloud, namely sparsity and varying density, which is also the key difference to indoor scenes with dense and uniform-density points. However, previous 3D voxelization methods consider the point cloud as a uniform one and split them via the uniform cube, while neglecting the varying-density property of outdoor point cloud. Consequently, this effect to apply the 3D partition to outdoor point cloud is met with fundamental difficulty.
   
   Motivated by these findings, we propose a new framework to outdoor LiDAR segmentation that consists of two key components, \ie, 3D cylindrical partition and asymmetrical 3D convolution networks, which maintain the 3D geometric information and handle these issues from partition and networks, respectively. Here, cylindrical partition resorts to the cylinder coordinates to divide the point cloud dynamically according to the distance (Regions that are far away from the origin have much sparse points, thus requiring a larger cell), which produces a more balanced point distribution (as shown in Fig.~\ref{fig:2d_3d_conv}); while asymmetrical 3D convolution networks strengthen the horizontal and vertical kernels to match the point distribution of objects in the driving scene and enhance the robustness to the sparsity. Moreover, voxel based methods might divide the points with different categories into the same cell and cell label encoding would inevitably cause the information loss (for LiDAR-based segmentation tasks). To alleviate the interference of lossy label encoding, a point-wise module is introduced to further refine the features obtained from voxel-based network. 
   Overall, the cooperation of these components well maintains the geometric relation and tackle the difficulty of outdoor point cloud, thus improving the effectiveness of 3D frameworks.

   Since the learned features from our model can be used for downstream tasks, we benchmark our model on a variety of LiDAR-based perception tasks such LiDAR-based semantic segmentation, panoptic segmentation and 3D detection. For semantic segmentation, we evaluate the proposed method on several large-scale outdoor datasets, including SemanticKITTI~\cite{behley2019semantickitti}, nuScenes~\cite{nuscenes} and A2D2~\cite{geyer2019a2d2}. Our method achieves the state-of-the-art on the leaderboard of SemanticKITTI (both single-scan and multi-scan challenges) and also outperforms the existing methods on nuScenes and A2D2 with a large margin. We also extend the proposed cylindrical partition and asymmetrical 3D convolution networks to LiDAR panoptic segmentation and LiDAR 3D detection. For panoptic segmentation and 3D detection, experimental results on SemanticKITTI and nuScenes, respectively, show its strong performance and good generalization capability.
   
   The contributions of this work mainly lie in three aspects:
   \begin{itemize}
   \item[(1)] We reposition the focus of outdoor LiDAR segmentation from 2D projection to 3D structure, and further investigate the inherent properties (difficulties) of outdoor point cloud.
    \item[(2)] We introduce a new framework to explore the 3D geometric pattern and tackle these difficulties caused by sparsity and varying density, through cylindrical partition and asymmetrical 3D convolution networks. 
    \item[(3)] The proposed method achieves the state of art on LiDAR-based semantic segmentation, LiDAR panoptic segmentation and LiDAR point cloud 3D detection, which also demonstrates its strong generalization capability.
   \end{itemize}

\section{Related Work}
   
   
    \begin{figure*}
    \centering
    \includegraphics[width=1.0\linewidth]{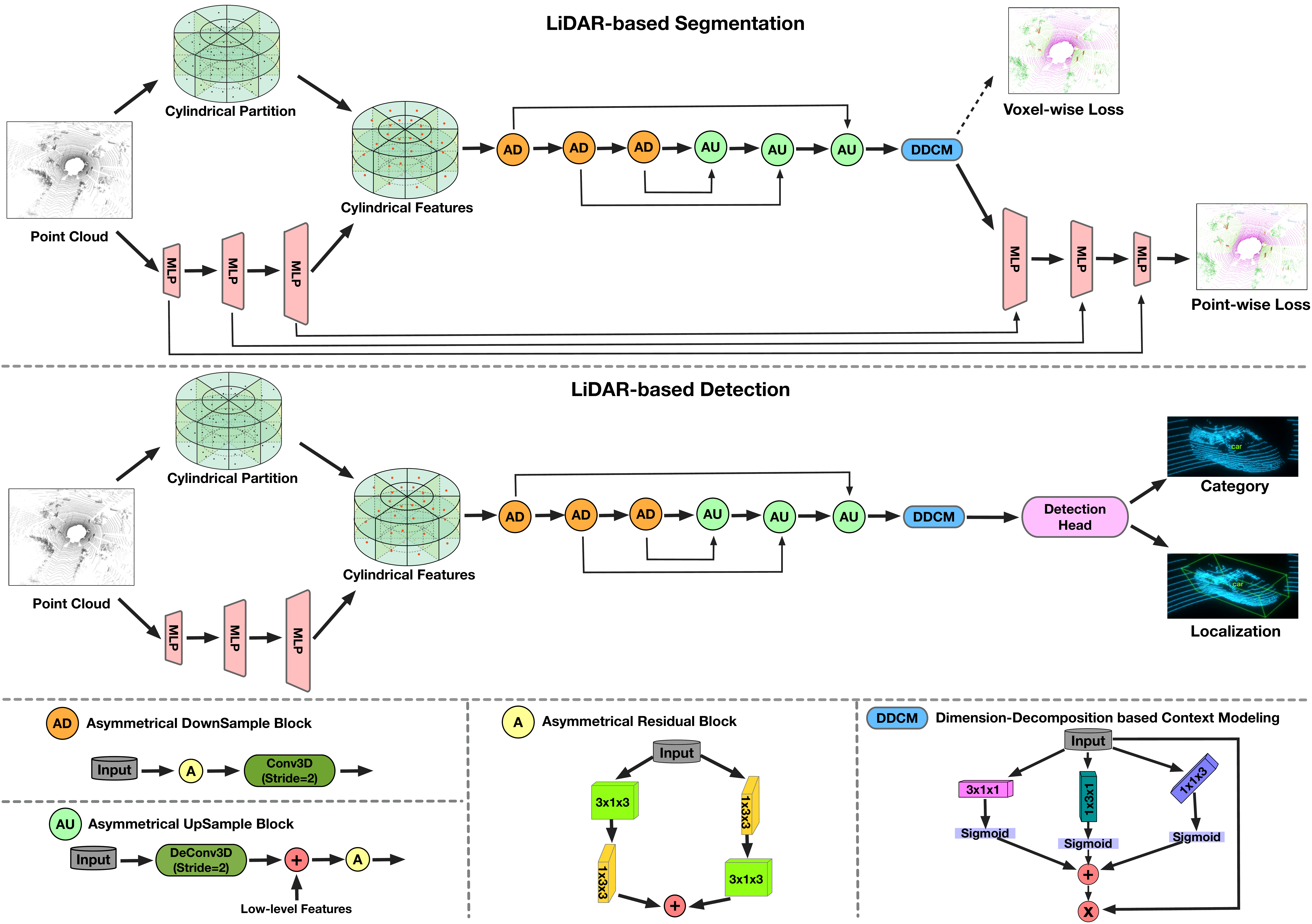}
    \caption{(1): Top row is the overall framework for LiDAR segmentation. Here, LiDAR point cloud is fed into MLP to get the point-wise features and then these features are reassigned based on the cylindrical partition. Asymmetrical 3D convolution networks are then used to generate the voxel-wise outputs. Finally, a point-wise module is introduced to refine these outputs.
    (2): Middle part shows the workflow of LiDAR 3D detection, where point-wise refinement is not attended.
    (3): Bottom row elaborates four components, including Asymmetrical Downsample block (AD), Asymmetrical Upsample blcok (AU), Asymmetrical residual block (A) and Dimension-Decomposition based Context Modeling (DDCM).}
    \label{fig:pipeline}
    \end{figure*}
   
   \myparagraph{Deep Learning for Indoor-scene Point Cloud.} Indoor-scene point clouds carry out some properties, including generally uniform density, small number of points, and small range of the scene. Mainstream methods~\cite{qi2017pointnet, thomas2019kpconv, wu2019pointconv, wang2019dynamic,velivckovic2017graph, lyu2020learning, engelmann20203d, zhang2020fusion, yan2020pointasnl,wang2019graph,pham2019jsis3d,qi20173d} of indoor point cloud segmentation learn the point features based on the raw point directly, which are often based on the pioneering work, \ie, PointNet, and promote the effectiveness of sampling, grouping and ordering to achieve the better performance. Another group of methods utilize the clustering algorithm~\cite{wang2019dynamic,velivckovic2017graph} to extract the hierarchical point features. However, these methods focusing on indoor point cloud are limited to adapt to the outdoor point cloud under the property of varying density and large range of scenes, and the large number of points also result in the computational difficulties for these methods when deploying from indoor to outdoor. 

   \myparagraph{Deep Learning for Outdoor-scene Point Cloud.}
   Most existing approaches for outdoor-scene point cloud~\cite{hu2020randla,cortinhal2020salsanext,zhao2021lif,milioto2019rangenet++,alonso20203d,zhang12356deep,landrieu2018large,hong2020lidar,cong2021input} focus on converting the 3D point cloud to 2D grids, to enable the usage of 2D Convolutional Neural Networks. SqueezeSeg~\cite{wu2018squeezeseg}, Darknet~\cite{behley2019semantickitti}, SqueezeSegv2~\cite{wu2019squeezesegv2},
  and RangeNet++~\cite{milioto2019rangenet++} utilize the spherical projection mechanism, which converts the point cloud to 
  a frontal-view image or a range image, and adopt the 2D convolution network on the pseudo image for point cloud segmentation or detection task. PolarNet~\cite{zhang2020polarnet} follows the bird's-eye-view projection, which projects point cloud data into bird's-eye-view representation under the polar coordinates. However, these 3D-to-2D projection methods inevitably loss and alter the 3D topology and fails to model the geometric information. Moreover, in most outdoor scenes, LiDAR device is often used to produce the point cloud data, where its inherent properties, \ie, sparsity and varying density , are often neglected.
  
  \myparagraph{3D Voxel Partition.}
  3D voxel partition is another routine of point cloud encoding~\cite{han2020occuseg,tchapmi2017segcloud,graham20183d,cciccek20163d,meng2019vv,cylinder3d1,cylinder3d2}. It converts a point cloud into 3D voxels, which mainly retains the 3D geometric information. 
  OccuSeg~\cite{han2020occuseg}, SSCN~\cite{graham20183d} and SEGCloud~\cite{tchapmi2017segcloud} follow this line to utilize the voxel partition and apply regular 3D convolutions for LiDAR segmentation. It is worth noting that while the aforementioned efforts have shown encouraging results, the improvement in the outdoor LiDAR point cloud remains limited. As mentioned above, a common issue is that these methods neglect the inherent properties of outdoor LiDAR point cloud, namely, sparsity and varying density. Compared to these methods, our proposed method resorts to the 3D cylindrical partition and asymmetrical 3D convolution networks to tackle these difficulties.

\myparagraph{Network Architectures for Feature Extraction}. Fully Convolutional Network~\cite{long2015fully} is the fundamental work for segmentation tasks in the deep-learning era. Built upon the FCN, many works aim to improve the performance via exploring the dilated convolution, multi-scale context modeling and attention modeling, including DeepLab\cite{chen2017deeplab, chen2018encoder} and PSP~\cite{zhao2017pyramid}. Recent work utilizes the neural architecture search to find the more effective backbone for the segmentation~\cite{liu2019auto,tang2020searching}. Particularly, U-Net~\cite{ronneberger2015u} proposes a symmetric architecture to incorporate the low-level features. With the great success of U-Net on 2D benchmarks and its good flexibility , many studies for LiDAR-based perception often adapt the U-Net to the 3D space~\cite{cciccek20163d}. We also follow this structure to construct our asymmetrical 3D convolution networks.


\section{Methodology}
\label{sec:methods}

    \begin{figure}[t]
    \centering
    \includegraphics[width=1.0\linewidth]{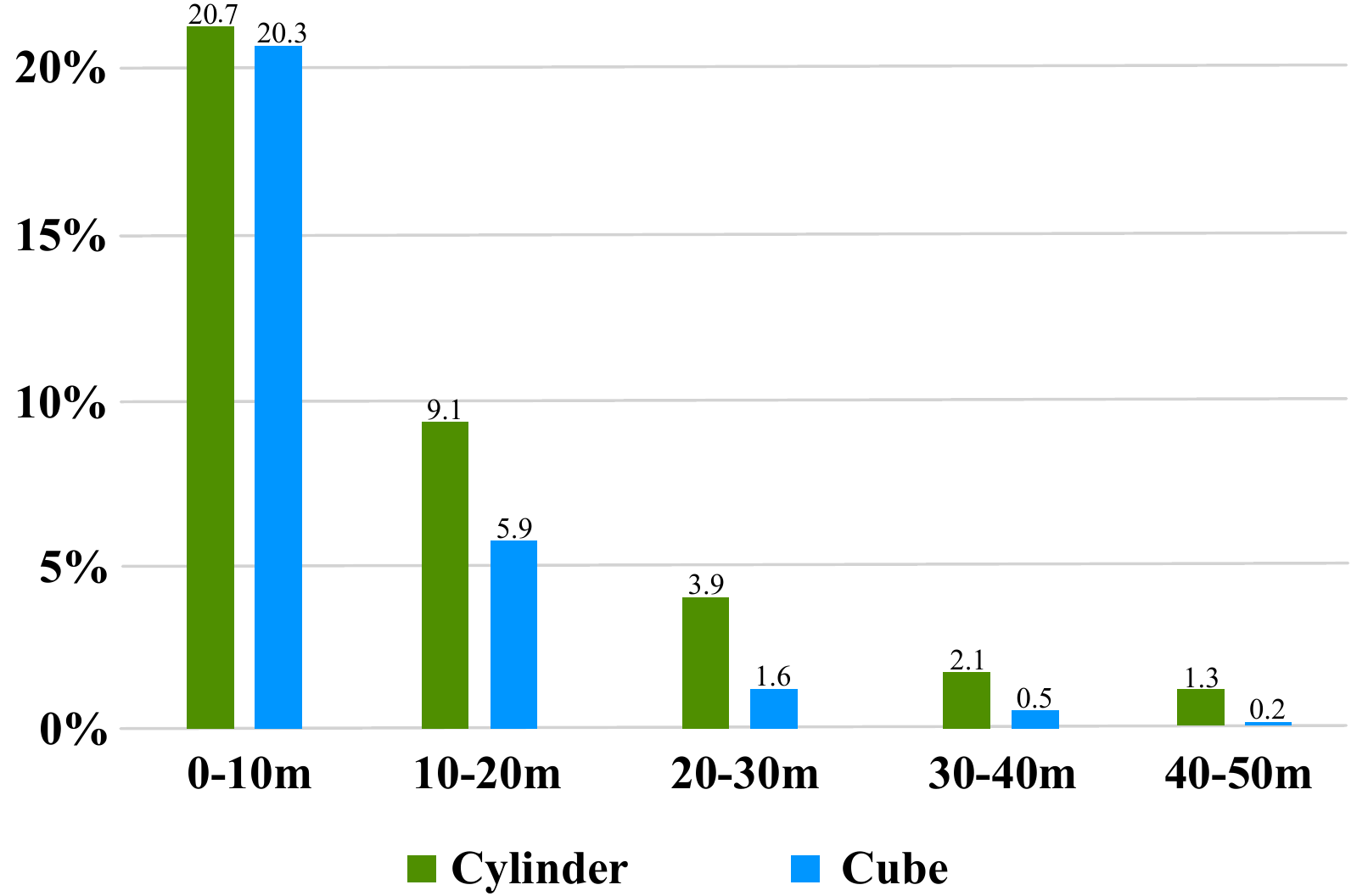}
    \caption{The proportion of non-empty cells at different distances between cylindrical and cubic partition (The results are calculated on the training set of SemanticKITTI). It can be found that cylinder partition makes a higher non-empty proportion and more balanced point distribution, especially for farther-away regions.}
    \label{fig:stat3}
    \end{figure}

    \begin{figure*}[t]
    \centering
    \includegraphics[width=1.0\linewidth]{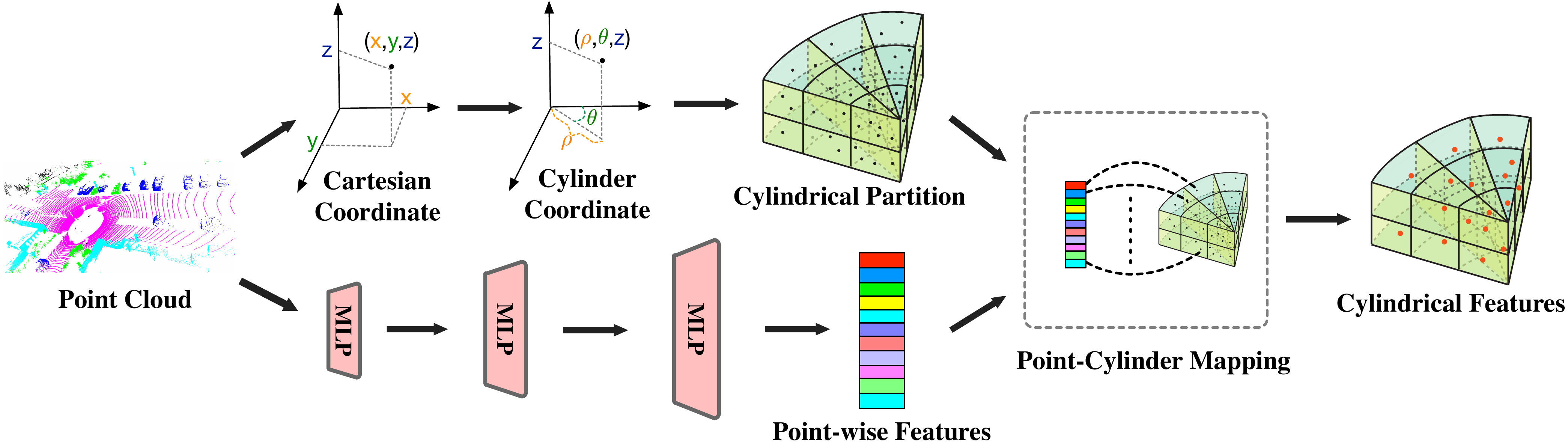}
    \caption{The pipeline of cylindrical partition. We first transform the Cartesian coordinate to Cylinder coordinate and then assign the point-wise features to the structured representation based on the Point-Cylinder mapping table. The cylindrical features are obtained via max-pooling these point-wise features inside each cylinder.}
    \label{fig:cylinder}
    \end{figure*}

\subsection{Framework Overview}

 As shown in the top and middle row of Fig.~\ref{fig:pipeline}, we elaborate the pipeline of our model in LiDAR-based segmentation and detection task.
In the context of semantic segmentation, given a point cloud, the task is to assign the semantic label to each point.
 Based on the comparison between 2D and 3D representation and investigation of the inherent properties of outdoor LiDAR point cloud, we desire to obtain a framework which explores the 3D geometric information and handles the difficulty caused by sparsity and varying-density. To this end, we propose a new outdoor segmentation approach based on the 3D partition and 3D convolution networks. To handle these difficulties of outdoor LiDAR point cloud, namely sparsity and varying density, we first employ the cylindrical partition to generate the more balanced point distribution (more robust to varying density), then apply the asymmetrical 3D convolution networks to power the horizontal and vertical weights, thus well matching the object point distribution in driving scene and enhancing the robustness to the sparsity. Same backbone with cylindrical partition and asymmetrical convolution network is also adapted to LiDAR-based 3D detection (shown in the middle row of Fig.~\ref{fig:pipeline}).

 
Specifically, the framework consists of two major components, including cylindrical partition and asymmetrical 3D convolution networks. The LiDAR point cloud is first divided by the cylindrical partition and the features extracted from MLP is then reassigned based on this partition. Asymmetrical 3D convolution networks are then used to generate the voxel-wise outputs. For segmentation tasks, a point-wise module is introduced to alleviate the interference of lossy cell-label encoding, thus refining the outputs. In the following sections, we will present these components in detail.

\subsection{Cylindrical Partition}

As mentioned above, outdoor-scene LiDAR point cloud possesses the property of varying density, where nearby region has much greater density than farther-away region. Therefore, uniform cells splitting the varying-density points would fall into an imbalanced distribution (for example, larger proportion of empty cells). While in the cylinder coordinate system, it utilizes the increasing grid size to cover the farther-away region, and thus more evenly distributes the points across different regions and gives an more balanced representation against the varying density. We perform a statistic to show the proportion of non-empty cells across different distances in Fig.~\ref{fig:stat3}. It can be found that with the distance goes far, cylindrical partition maintains a balanced non-empty proportion due to the increasing grid size while cubic partition suffers the imbalanced distribution, especially in the farther-away regions (about 6 times less than cylindrical partition). Moreover, unlike these projection-based methods project the point to the 2D view, cylindrical partition maintains the 3D grid representation to retain the geometric structure. 

The workflow is illustrated in Fig.~\ref{fig:cylinder}. We first transform the points on Cartesian coordinate system to the Cylinder coordinate system. This step transforms the points ($x, y, z$) to points ($\rho, \theta, z$), where radius $\rho$ (distance to origin in x-y axis) and azimuth $\theta$ (angle from x-axis to y-axis) are calculated. Then cylindrical partition performs the split on these three dimensions, note that in the cylinder coordinate, the farther-away the region is, the larger the cell will be.
Point-wise features obtained from the MLP are reassigned based on the result of this partition to get the cylindrical features.
Specifically, the point-cylinder mapping contains the index of point-wise features to cylinder. Based on this mapping function, point-wise features within same cylinder are mapped together and processed via max-pooling to get the cylindrical features.
After these steps, we unroll the cylinder from 0-degree and get the 3D cylindrical representation $\mathbb{F}\in C \times H\times W \times L$, where $C$ denotes the feature dimension and $H, W, L$ mean the radius, azimuth and height.
Subsequent asymmetrical 3D convolution networks will be performing on this representation.

    \begin{figure*}[t]
    \centering
    \includegraphics[width=1.0\linewidth]{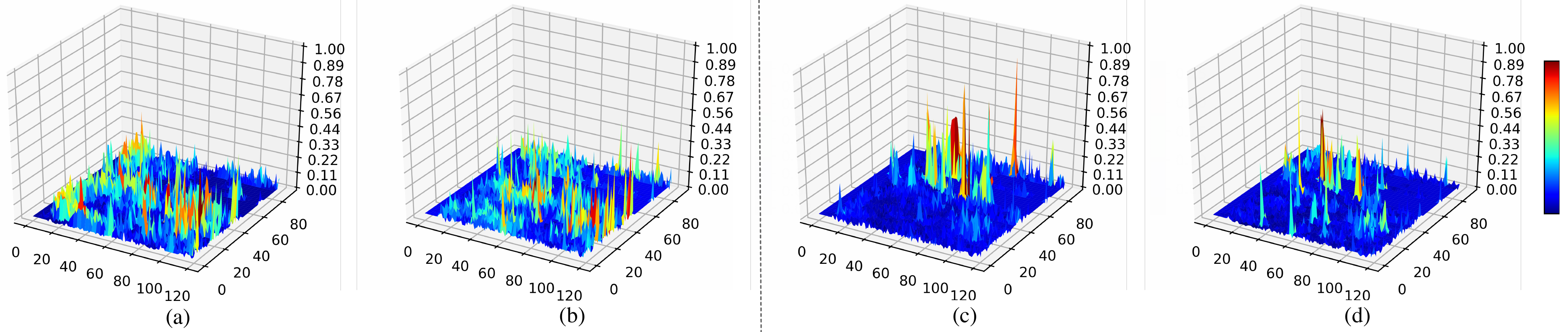}
    \caption{Filter activation visualization. We visualize some filters at second downsampling block from regular 3D convolution networks (with cubic partition) (Fig.5 (a) and (b)) and asymmetrical 3D convolution networks (with cylindrical partition) (Fig.5 (c) and (d)), respectively. Since each filter in 3D convolution networks has size of H$\times$W$\times$L, we use the mean value of L dimension as the activation value. Fig (a) and (c) are extracted from 100th filter and Fig (b) and (d) are extracted from 120th filter. It can be observed that the filters from the proposed asymmetrical 3D convolution networks are sparsely activated in some certain regions while the regular convolution covers most parts.}
    \label{fig:acti}
    \end{figure*}

\subsection{Asymmetrical 3D Convolution Network}

Since the driving-scene point cloud carries out the specific object shape distribution, including car, truck, bus, motorcycle and other cubic objects, we aim to follow this observation to enhance the representational power of a standard 3D convolution. Moreover, recent literature~\cite{wang2019shape,ding2019acnet} also shows that the central crisscross weights count more in the square convolution kernel. In this way, we devise the asymmetrical residual block to strengthen the horizontal and vertical responses and match the object point distribution. 
Based on the proposed asymmetrical residual block, we further build the asymmetrical downsample block and asymmetrical upsample block to perform the downsample and upsample operation. Moreover, a dimension-decomposition based context modeling (termed as DDCM) is introduced to explore the high-rank global context in decomposite-aggregate strategy. We detail these components in the bottom of Fig.~\ref{fig:pipeline}


    \begin{figure}
    \centering
    \begin{subfigure}[t]{0.45\linewidth}
    \centering
\includegraphics[width=1\linewidth]{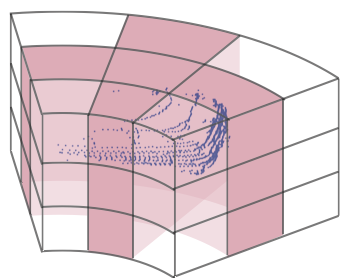}
    \caption{Car}
    \label{fig:asym1}
    \end{subfigure}
    \begin{subfigure}[t]{0.45\linewidth}        \includegraphics[width=1\linewidth]{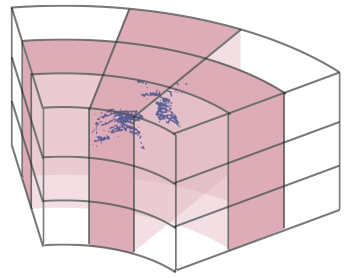}
    \caption{Motorcycle}
    \label{fig:asym2}
    \end{subfigure}
    \vspace{-1ex}
\caption{An illustration of asymmetrical residual block, where two asymmetrical kernels are stacked to power the skeleton. It can be observed that asymmetrical residual block focuses on the horizontal and vertical kernels.}
    \label{fig:asym}
\end{figure}

\myparagraph{Asymmetrical Residual Block}
Motivated by the observation and conclusion in~\cite{wang2019shape,ding2019acnet}, the asymmetrical residual block strengthens the horizontal and vertical kernels, which matches the point distribution of object in the driving scene and explicitly makes the skeleton of the kernel powerful, thus enhancing the robustness to the sparsity of outdoor LiDAR point cloud. We use the Car and Motorcycle as the example to show the asymmetrical residual block in Fig.~\ref{fig:asym}, where 3D convolutions are performing on the cylindrical grids. Moreover, the proposed asymmetrical residual block also saves the computation and memory cost compared to the regular square-kernel 3D convolution block. By incorporating the asymmetrical residual block, the asymmetrical downsample block and upsample block are designed and our asymmetrical 3D convolution networks are built via stacking these downsample and upsample blocks.

\myparagraph{Dimension-Decomposition based Context Modeling}
Since the global context features should be high-rank to have enough capacity to capture the large context varieties~\cite{zhang2019co}, it is hard to construct these features directly. We follow the tensor decomposition theory~\cite{chen2020tensor} to build the high-rank context as a combination of low-rank tensors, where we use three rank-1 kernels to obtain the low-rank features and then aggregate them together to get the final global context.

\begin{figure}[t]
    \centering
    \includegraphics[width=1.0\linewidth]{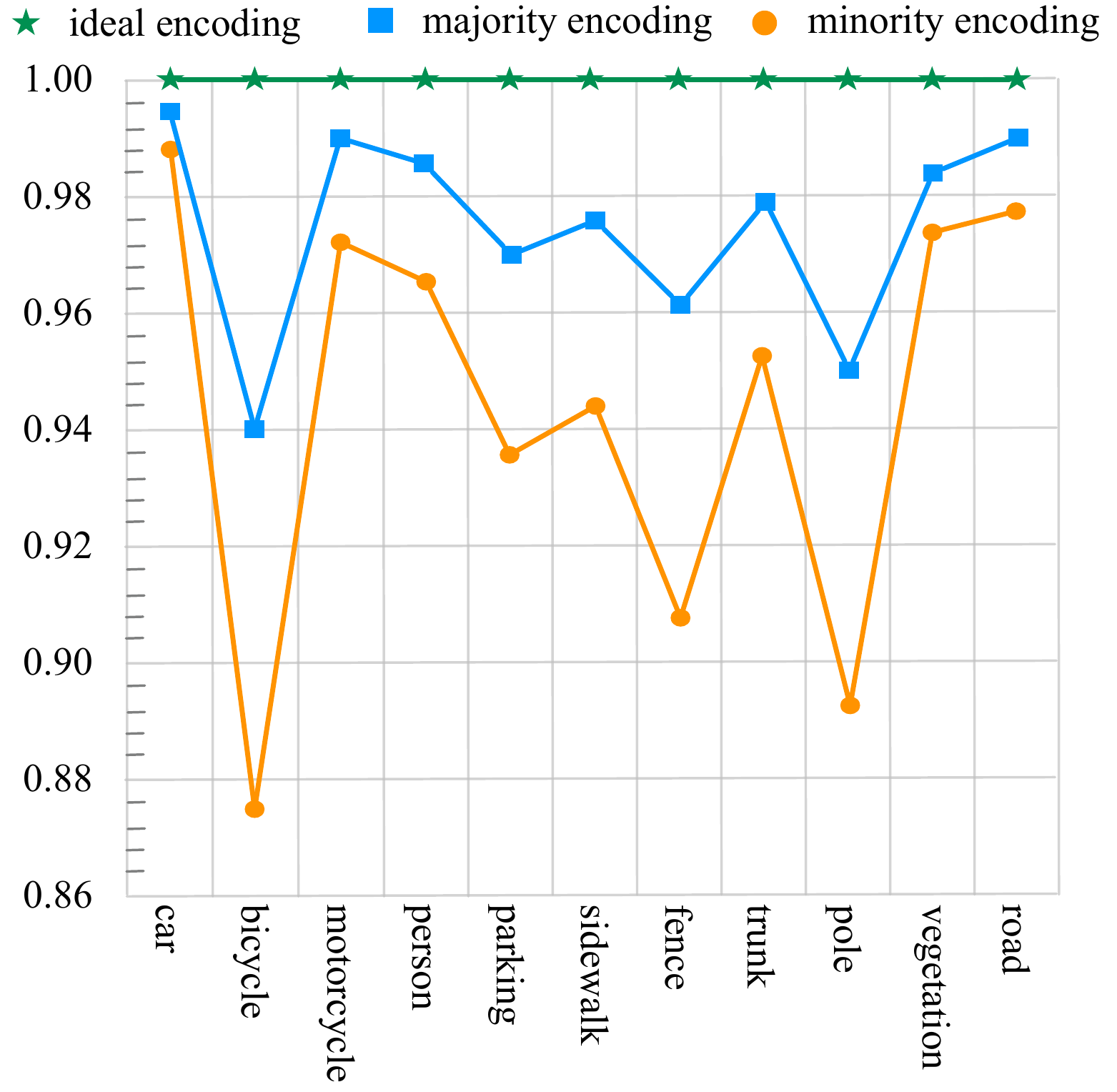}
    \caption{Upper bound of mIoU with different label encoding methods (\ie, majority and minority encoding). It can be found that no matter what encoding methods are, the information loss always occurs, which is also the reason for point-wise refinement. }
    \label{fig:stat2}
\end{figure}

\subsection{Sparse Activation Visualization}

As mentioned above, the proposed cylindrical partition and asymmetrical 3D networks aim to tackle the difficulties caused by sparsity and varying-density in outdoor point cloud. We thus visualize some filter activations from regular 3D convolution networks (with regular cubic partition) and asymmetrical 3D convolution networks (with cylindrical partition), respectively. The results are shown in Fig.~\ref{fig:acti}. Fig.~\ref{fig:acti}(a) and (b) are extracted from regular 3D convolution networks, which are activated at almost regions; While the proposed asymmetrical 3D convolution networks strengthen sparser activations and focus on them (as shown in Fig.~\ref{fig:acti}(c) and (d)), they mainly focus on some certain regions. It demonstrates that the proposed model could adaptively handle the sparse point cloud input and focus on some certain regions.

\subsection{Point-wise Refinement Module}

\begin{figure}[t]
    \centering
    \includegraphics[width=0.8\linewidth]{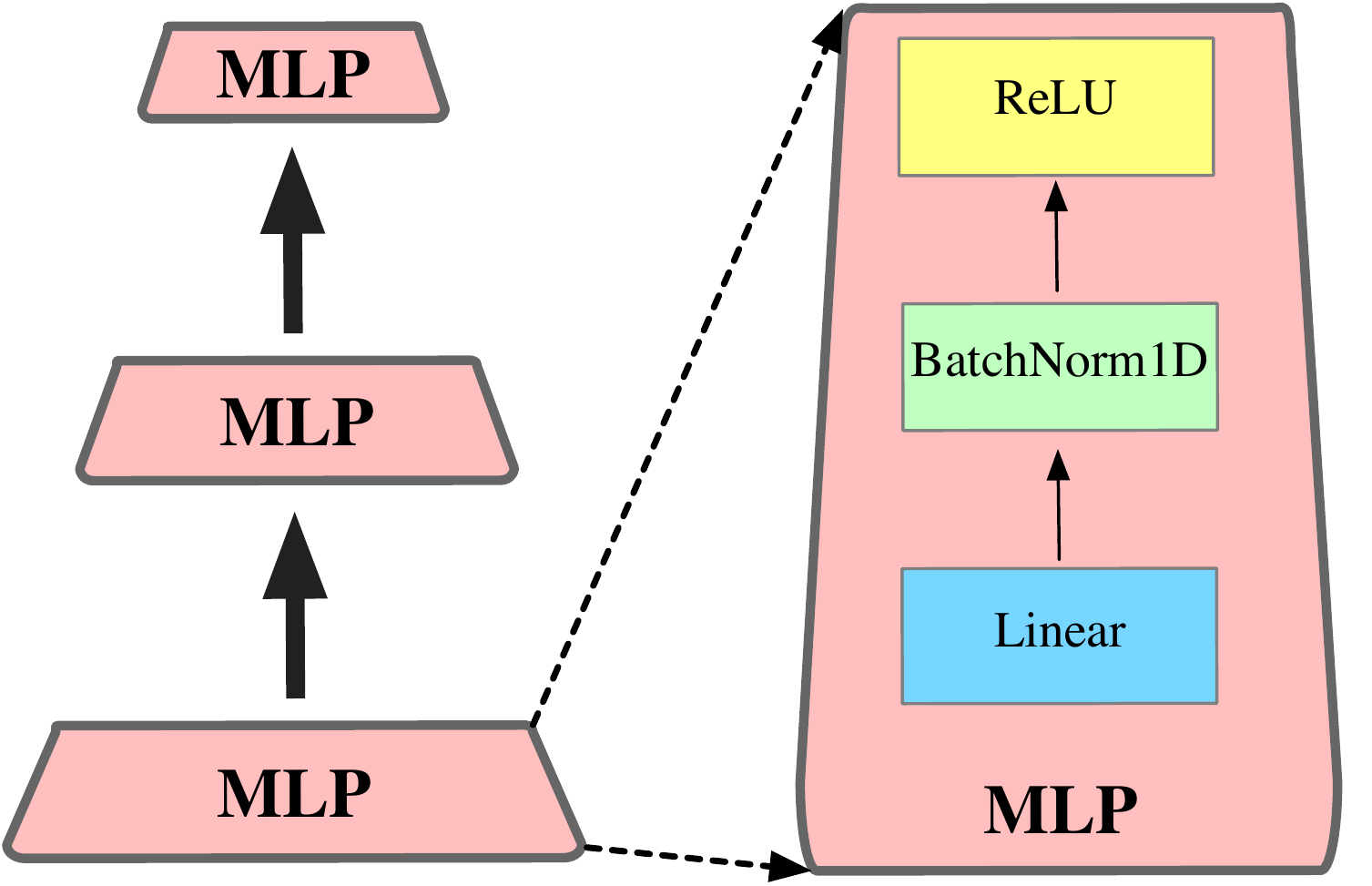}
    \caption{Detailed workflow of MLPs in Cylindrical Partition and Point-wise Refinement Module.}
    \label{fig:prnet}
\end{figure}

Partition-based methods predict one label for each cell. Although partition-based methods effectively explore the large-range point cloud, however, this group of method, including cube-based and cylinder-based, inevitably suffers from the lossy cell-label encoding, \eg, points with different categories are divided into same cell, and this case would cause the information loss for point cloud semantic segmentation task (as shown in the middle row of Fig.~\ref{fig:pipeline}). 
We make a statistic to show the effect of different label encoding methods with cylindrical partition in Fig.~\ref{fig:stat2}, where \emph{majority encoding} means using the major category of points inside a cell as the cell label and \emph{minority encoding} indicates using the minor category as the cell label. It can be observed that both of them cannot reach the 100 percent mIoU (ideal encoding) and inevitably have the information loss. Here, the point-wise refinement module is introduced to alleviate the interference of lossy cell-label encoding. We first project the cylindrical features to the point-wise based on the inverse point-cylinder mapping table (note that points inside same cylinder would be assigned to the same cylindrical features). Then the point-wise module takes both point features before and after 3D convolution networks as the input, and fuses them together to refine the output. We also show the detailed structure of MLPs in point-wise refinement module and cylindrical partition in Fig.~\ref{fig:prnet}.

\subsection{Objective Function}

For LiDAR-based semantic segmentation task, the total objective of our method consists of two components, including voxel-wise loss and point-wise loss. It can be formulated as $\mathbb{L} = \mathbb{L}_{voxel} + \mathbb{L}_{point}$.
For the voxel-wise loss ($\mathbb{L}_{voxel}$), we follow the existing methods~\cite{cortinhal2020salsanext,hu2020randla} and use the weighted cross-entropy loss and lovasz-softmax~\cite{berman2018lovasz} loss to maximize the point accuracy and the intersection-over-union score, respectively. For point-wise loss ($\mathbb{L}_{point}$), we only use the weighted cross-entropy loss to supervise the training. During inference, the output from point-wise refinement module is used as the final output. 

For LiDAR-based panoptic segmentation task, except the loss of semantic segmentation, it also contains the loss of instance branch~\cite{hong2020lidar}, which utilizes center regression to achieve the clustering.

For LiDAR-based 3D detection, we follow previous work~\cite{zhu2020ssn,yan2018second} to use the focal loss and smooth L1 loss for classification and regression, respectively.


\section{Experiments}

In this section, we benchmark the proposed model on three downstream tasks. For semantic segmentation task, we evaluate the proposed method on several large-scale datasets, \ie, SemanticKITTI, nuScenes and A2D2. SemanticKITTI and nuScenes are also used in panoptic segmentation and 3D detection, respectively.
Furthermore, extensive ablation studies on LiDAR semantic segmentation task are conducted to validate each component. 

\begin{table*}[t]
\caption{Results of our proposed method and state-of-the-art LiDAR Segmentation methods on SemanticKITTI single-scan test set. Note that all results are obtained from the literature, where post-processing, flip \& rotation test ensemble, \etc are also applied.}
\label{semantickitti}
\centering
\begin{adjustbox}{width=\textwidth}
\begin{tabular}{c|c|c|c|c|c|c|c|c|c|c|c|c|c|c|c|c|c|c|c|c}
\hline
\textbf{Methods} & \textbf{mIoU} & \rotatebox{90}{car} &  \rotatebox{90}{bicycle} & \rotatebox{90}{motorcycle} & \rotatebox{90}{truck} & \rotatebox{90}{other-vehicle} & \rotatebox{90}{person} & \rotatebox{90}{bicyclist} & \rotatebox{90}{motorcyclist} & \rotatebox{90}{road} & \rotatebox{90}{parking} & \rotatebox{90}{sidewalk} & \rotatebox{90}{other-ground} &
\rotatebox{90}{building} & \rotatebox{90}{fence} & \rotatebox{90}{vegetation} & \rotatebox{90}{trunk} & \rotatebox{90}{terrain} & \rotatebox{90}{pole} & \rotatebox{90}{traffic} \\
\hline
\hline
TangentConv~\cite{tatarchenko2018tangent} & 35.9 & 86.8 & 1.3 & 12.7 & 11.6 & 10.2 & 17.1 & 20.2 & 0.5 & 82.9 & 15.2 & 61.7 & 9.0 & 82.8 & 44.2 & 75.5 & 42.5 & 55.5 & 30.2 & 22.2\\
\hline
Darknet53~\cite{behley2019semantickitti} & 49.9 & 86.4 & 24.5 & 32.7 & 25.5 & 22.6 & 36.2 & 33.6 & 4.7 & 91.8 & 64.8 & 74.6 & {27.9} & 84.1 & 55.0 & 78.3 & 50.1 & 64.0 & 38.9 & 52.2 \\
\hline
RandLA-Net~\cite{hu2020randla} & 50.3 & 94.0 & 19.8 & 21.4 & {42.7} & 38.7 & 47.5 & 48.8 & 4.6  & 90.4 & 56.9 & 67.9 & 15.5 & 81.1 & 49.7 & 78.3 & 60.3 & 59.0 & 44.2 & 38.1 \\
\hline
RangeNet++~\cite{milioto2019rangenet++} & 52.2 & 91.4 & 25.7 & 34.4 & 25.7 & 23.0 & 38.3 &  38.8 & 4.8 & {91.8} & {65.0} & 75.2 & 27.8 & 87.4 & 58.6 & 80.5 & 55.1 & 64.6 & 47.9 & 55.9 \\
\hline
PolarNet~\cite{zhang2020polarnet} & 54.3 & 93.8 & 40.3 & 30.1 & 22.9 & 28.5 & 43.2 & 40.2 & 5.6 & 90.8 & 61.7 & 74.4 & 21.7 & {90.0} & 61.3 & 84.0 & 65.5 & 67.8 & 51.8 & 57.5  \\
\hline
SqueezeSegv3~\cite{xu2020squeezesegv3} & 55.9 & 92.5 & 38.7 & 36.5 & 29.6 & 33.0 & 45.6 & 46.2 & {20.1} & 91.7 & 63.4 & 74.8 & 26.4 & 89.0 & 59.4 & 82.0 & 58.7 & 65.4 & 49.6 & 58.9  \\
\hline
Salsanext~\cite{cortinhal2020salsanext} & 59.5 & 91.9 & 48.3 & 38.6 & 38.9 & 31.9 & 60.2 & 59.0 & 19.4 & 91.7 & 63.7 & 75.8 & 29.1 & 90.2 & 64.2 & 81.8 & 63.6 & 66.5 & 54.3 & 62.1 \\
\hline
KPConv~\cite{thomas2019kpconv} &58.8& 96.0&32.0 & 42.5 & 33.4&44.3&61.5 & 61.6 & 11.8 & 88.8 & 61.3&  72.7&31.6& \bf{95.0} & 64.2 & 84.8 & 69.2 & 69.1 & 56.4 & 47.4 \\
\hline
FusionNet~\cite{zhang12356deep} & 61.3 & 95.3 & 47.5 & 37.7 & 41.8 & 34.5 & 59.5 & 56.8 & 11.9 & 91.8 & 68.8 & 77.1 & 30.8 & 92.5 & \bf{69.4} & 84.5 & 69.8 & 68.5&60.4 & \bf{66.5} \\ 
\hline
KPRNet~\cite{kochanov2020kprnet} & 63.1 & 95.5&54.1& 47.9&23.6 & 42.6&65.9 & 65.0 & 16.5 & \bf{93.2} & \bf{73.9} & \bf{80.6} & 30.2 & 91.7 & {68.4} & \bf{85.7} & 69.8 & 71.2 & 58.7 & 64.1 \\
\hline
TORANDONet~\cite{gerdzhev2020tornado} & 63.1 &94.2& 55.7& 48.1& 40.0& 38.2& 63.6& 60.1& 34.9& 89.7& 66.3& 74.5& 28.7& 91.3& 65.6& 85.6& 67.0& \bf{71.5} & 58.0 & {65.9} \\
\hline
SPVNAS~\cite{tang2020searching} & 66.4 & - & - & - & - & - & - & - & - & - & - & - & - & - & - & - & - & - & - & - \\
\hline
\hline
Ours & \bf{67.8} & \bf{97.1} & \bf{67.6} & \bf{64.0} & \textbf{59.0} & \bf{58.6} & \bf{73.9} & \bf{67.9} & \bf{36.0} & {91.4} & {65.1} & {75.5} & \bf{32.3} & {91.0} & {66.5} & {85.4} & \bf{71.8} & {68.5} & \bf{62.6} & {65.6}  \\
 \hline
\end{tabular}
\end{adjustbox}
\end{table*}

\begin{table*}[t]
\caption{Results of our proposed method and other LiDAR Segmentation methods on nuScenes validation set.}
\label{nuscenes}
\centering
\begin{adjustbox}{width=\textwidth}
\begin{tabular}{c|c|c|c|c|c|c|c|c|c|c|c|c|c|c|c|c|c}
\hline
\textbf{Methods} & \textbf{mIoU} & \rotatebox{90}{barrier} &  \rotatebox{90}{bicycle} & \rotatebox{90}{bus} & \rotatebox{90}{car} & \rotatebox{90}{construction} & \rotatebox{90}{motorcycle} & \rotatebox{90}{pedestrian} & \rotatebox{90}{traffic-cone} & \rotatebox{90}{trailer} & \rotatebox{90}{truck} & \rotatebox{90}{driveable} & \rotatebox{90}{other} &
\rotatebox{90}{sidewalk} & \rotatebox{90}{terrain} & \rotatebox{90}{manmade} & \rotatebox{90}{vegetation} \\
\hline
\hline
RangeNet++~\cite{milioto2019rangenet++} & 65.5 & 66.0 & 21.3 & 77.2 & 80.9 & 30.2 & 66.8 & 69.6 &  52.1 & 54.2 & {72.3} & {94.1} & 66.6 & 63.5 & 70.1 & 83.1 & 79.8 \\
\hline
PolarNet~\cite{zhang2020polarnet} & 71.0 & 74.7 & 28.2 & 85.3 & 90.9 & 35.1 & 77.5 & 71.3 & 58.8 & 57.4 & 76.1 & 96.5 & 71.1 & 74.7 & {74.0} & 87.3 & 85.7  \\
\hline
Salsanext~\cite{cortinhal2020salsanext} & 72.2 & 74.8 & 34.1 & 85.9 & 88.4 & 42.2 & 72.4 & 72.2 & 63.1 & 61.3 & 76.5 & 96.0 & 70.8 & 71.2 & 71.5 & 86.7 & 84.4 \\
\hline
\hline
Ours & \bf{76.1} & \bf{76.4} & \bf{40.3} & \bf{91.2} & \bf{93.8} & \textbf{51.3} & \bf{78.0} & \bf{78.9} & \bf{64.9} & \bf{62.1} & \bf{84.4} & \bf{96.8} & \bf{71.6} & \bf{76.4} & \bf{75.4} & \bf{90.5} & \bf{87.4}  \\
 \hline
\end{tabular}
\end{adjustbox}
\end{table*}

\subsection{Dataset and Metric}

\myparagraph{SemanticKITTI~\cite{behley2019semantickitti}}~~~ is a large-scale driving-scene dataset for point cloud segmentation, including semantic segmentation and panoptic segmentation. It is derived from the KITTI Vision Odometry Benchmark and collected in Germany with the Velodyne-HDLE64 LiDAR. The dataset consists of 22 sequences, splitting sequences 00 to 10 as training set (where sequence 08 is used as the validation set), and sequences 11 to 21 as test set. 19 classes are remained for training and evaluation after merging classes with different moving status and ignore classes with very few points. In this dataset, it consists of two challenges, namely, single-scan and multi-scan point-cloud semantic segmentation, where single-scan denotes the single-frame point cloud semantic segmentation and multi-scan denotes the multiple-frame point cloud segmentation, respectively. The key difference is that multi-scan semantic segmentation requires classifying the moving categories, including moving car, moving truck, moving person, moving bicyclist, moving motorcyclist.

\myparagraph{nuScenes~\cite{nuscenes}} It collects 1000 scenes of 20s duration with 32 beams LiDAR sensor. The number of total frames is 40,000, which is sampled at 20Hz. They also officially split the data into training and validation set. After merging similar classes and removing rare classes, total 16 classes for the LiDAR semantic segmentation are remained.

\myparagraph{A2D2~\cite{geyer2019a2d2}} We follow the data pre-processing in~\cite{zhang2020polarnet} to generate the label and process the point cloud data. A2D2 uses five asynchronous LiDAR sensors where each sensor covers a potion of the surrounding view. After LiDAR panoramic stitching, the A2D2 dataset is split into 22408, 2774 and 13264 training, validation and test scans, respectively with 38-class segmentation annotation. Since there are 38 categories in A2D2 dataset where some of them only have subtle differences, it is harder than other datasets, SemanticKITTI and nuScenes. 

\begin{figure*}[t]
\centering
\begin{tabular}{c}
\includegraphics[width=0.8\linewidth]{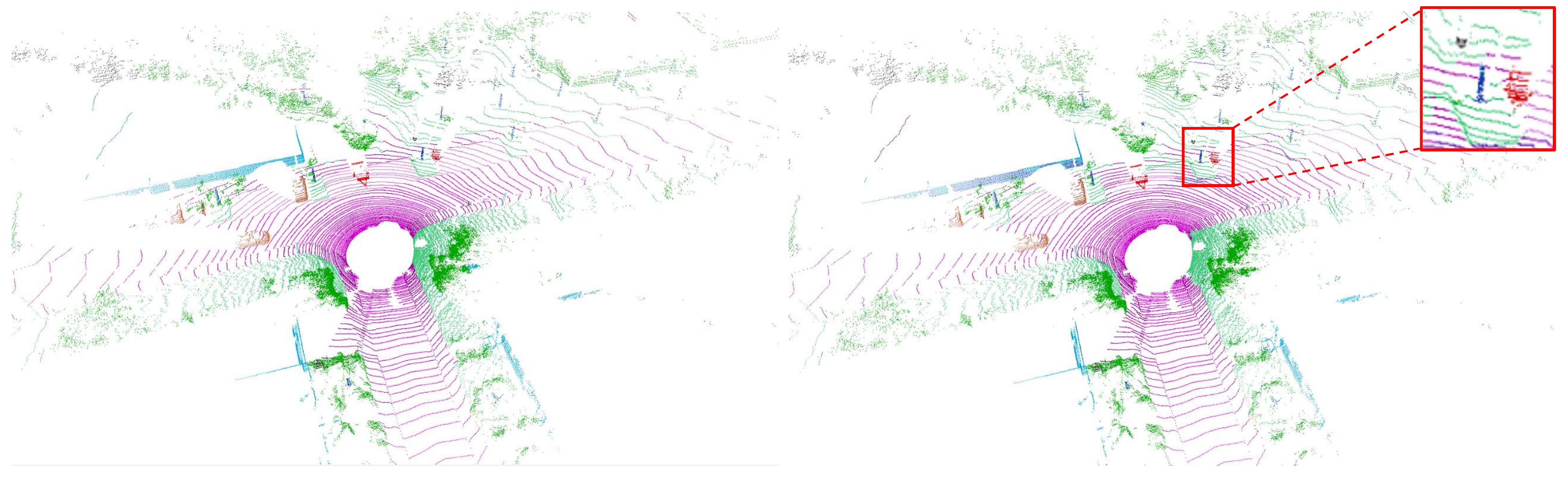} \\
\hline
\includegraphics[width=0.8\linewidth]{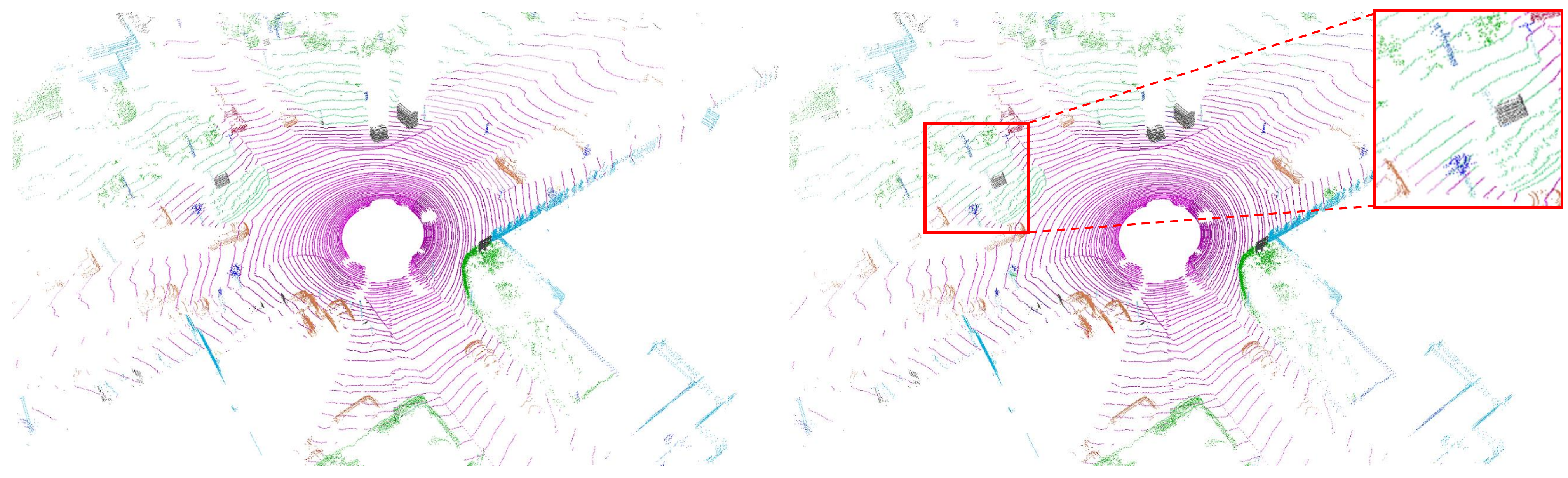} \\
\end{tabular}
\caption{Visualization of single-scan semantic segmentation on SemanticKITTI validation set. The left is ground-truth and right is our prediction.}
\label{fig:vis_result}
\end{figure*}

\myparagraph{Implementation Details}
For these datasets, the Cartesian spaces are different which are related to the LiDAR sensor range. In our implementation, we fix the Cartesian spaces to be $[x: \pm 50m, y: \pm 50m, z: -3m \sim 1.5m]$, $[x: \pm 50m, y: \pm 50m, z: -4m \sim 2m]$, and $[x: \pm 50m, y: \pm 50m, z: -3m \sim 9m]$ for SemanticKITTI, nuScenes and A2D2, respectively. After transforming to the Cylindrical spaces, they are fixed to be $[\rho: 0m \sim 50m, \theta: \pm \pi, z: -3m \sim 1.5m]$, $[\rho: 0m \sim 50m, \theta: \pm \pi, z: -4m \sim 2m]$, and $[\rho: 0m \sim 50m, \theta: \pm \pi, z: -9m \sim 3m]$. In this way, the proposed cylindrical spaces can cover more than 99\% of points for each point cloud scan on average and points out of the spaces are assigned to the closest cylindrical cell.
For all datasets, cylindrical partition splits these point clouds into 3D representation with the size = $480\times360\times32$, where three dimensions indicate the radius, angle and height, respectively. We also perform the ablation studies to investigate and cross-validate the effect of these parameters $480\times360\times32$. We use NVIDIA V100 GPU with 16G memory to train the proposed model with batch size = 2.

\myparagraph{Evaluation Metric}
To evaluate the proposed method, we follow the official guidance to leverage mean intersection-over-union (mIoU) as the evaluation metric defined in~\cite{behley2019semantickitti,nuscenes}, which can be formulated as:
$
IoU_{i} = \frac{TP_{i}}{TP_{i} + FP_{i} + FN_{i}}
$
where $TP_{i}, FP_{i}, FN_{i}$ represent true positive, false positive, and false negative predictions for
class $i$ and the mIoU is the mean value of $IoU_{i}$ over all classes.

\subsection{LiDAR-based Semantic Segmentation}

\myparagraph{Results on SemanticKITTI Single-scan Semantic Segmentation}
In this experiment, we compare the results of our proposed method with existing state-of-the-art LiDAR segmentation methods on SemanticKITTI single-scan test set. The target is to generate the semantic prediction for single frame point cloud.
As shown in Table~\ref{semantickitti}, our method outperforms all existing methods in term of mIoU. Compared to the projection-based methods on 2D space, including Darknet53~\cite{behley2019semantickitti}, SqueezeSegv3~\cite{xu2020squeezesegv3}, RangeNet++~\cite{milioto2019rangenet++} and PolarNet~\cite{zhang2020polarnet}, our method achieves 8\% $\thicksim$ 17\% performance gain in term of mIoU due to the modeling of 3D geometric information. Compared to some voxel partition and 3D convolution based methods, including FusionNet~\cite{zhang12356deep}, TORANDONet~\cite{gerdzhev2020tornado} (multi-view fusion based method) and SPVNAS~\cite{tang2020searching} (utilizing the neural architecture search for LiDAR segmentation), the proposed method also performs better than these 3D convolution based methods, where the cylindrical partition and asymmetrical 3D convolution networks well handle the difficulty of driving-scene LiDAR point cloud that is neglected by these methods.

\myparagraph{Visualization} We show some visualization results of single-scan segmentation in Fig.\ref{fig:vis_result}, which are sampled from the SemanticKITTI validation set.
It can be observed that the proposed method mainly achieves decent accuracy, and well separates the nearby objects and accurately identifies them because it maintains the 3D topology and utilizes the geometric information 
(we highlight corresponding regions with red rectangles). These visualization can verify our claim that keeping 3D structure and more balanced point distribution could benefit the segmentation results.

\begin{figure}[t]
    \centering
\includegraphics[width=0.99\linewidth]{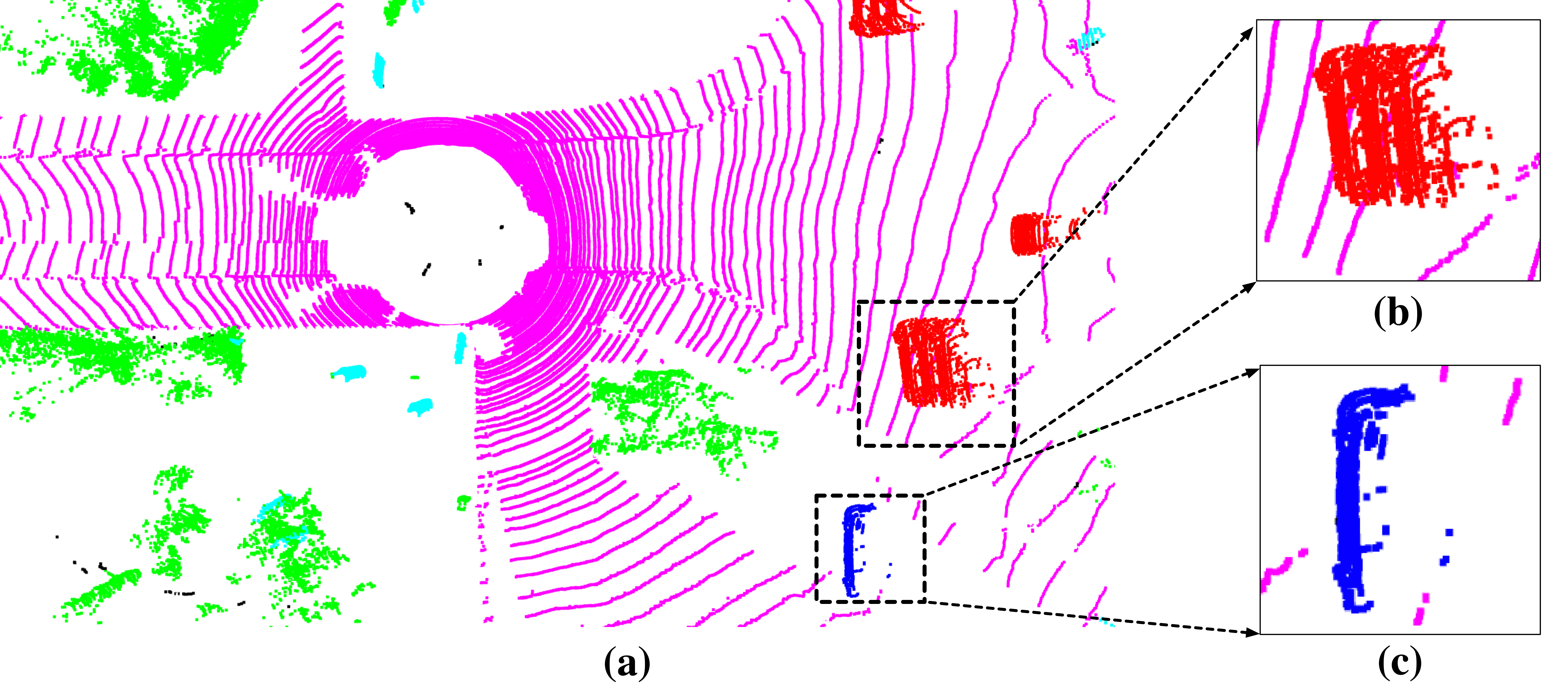}
    \caption{An example of multi-scan fusion. (b) and (c) represent the moving car and stationary car after the multi-scan fusion, respectively. Note that we use three frames to perform the multi-scan fusion.}
    \label{fig:multiscan}
\end{figure}

\begin{table*}[t]
\caption{Results of our proposed method and state-of-the-art LiDAR Segmentation methods on SemanticKITTI multi-scan test set - Part \uppercase\expandafter{\romannumeral1}.}
\label{semantickitti-multiscan}
\centering
\begin{adjustbox}{width=\textwidth}
\begin{tabular}{c|c|c|c|c|c|c|c|c|c|c|c|c|c|c|c}
\hline
\textbf{Methods} & \textbf{mIoU} & \rotatebox{90}{car} &  \rotatebox{90}{bicycle} & \rotatebox{90}{motorcycle} & \rotatebox{90}{truck} & \rotatebox{90}{other-vehicle} & \rotatebox{90}{person} & \rotatebox{90}{bicyclist} & \rotatebox{90}{motorcyclist} & \rotatebox{90}{road} & \rotatebox{90}{parking} & \rotatebox{90}{sidewalk} & \rotatebox{90}{other-ground} &
\rotatebox{90}{building} & \rotatebox{90}{fence} \\
\hline
\hline
TangentConv~\cite{tatarchenko2018tangent} & 34.1 & 84.9 & 2.0 & 18.2 & 21.1 & 18.5 & 1.6 & 0.0 & 0.0 & 83.9 & 38.3 & 64.0 & 15.3 & 85.8 & {49.1}  \\
 \hline
DarkNet53~\cite{semantickitti} & 41.6 & 84.1 & 30.4 & 32.9 & 20.0 & 20.7 & 7.5 & 0.0&0.0 & \textbf{91.6} & 64.9 & \bf{75.3} & 27.5 & 85.2 & {56.5}  \\
 \hline
SpSeqnet~\cite{shi2020spsequencenet} & 43.1 & 88.5 & 24.0 & 26.2 & 29.2 & 22.7 & 6.3 & 0.0 & 0.0 & 90.1 & 57.6 & 73.9 & 27.1 & 91.2 & \bf{66.8}  \\ 
\hline
KPConv~\cite{thomas2019kpconv} &51.2& 93.7&44.9 & 47.2 & \textbf{42.5}& 38.6 & \textbf{21.6} & 0.0 & 0.0 & 86.5 & 58.4&  70.5&26.7& {90.8} & 64.5 \\
\hline
\hline
Ours & \bf{51.5} & \bf{93.8} & \bf{67.6} & \bf{63.3} & {41.2} & \bf{37.6} & {12.9} & \bf{0.1} & \bf{0.1} & {90.4} & \bf{66.3} & {74.9} & \bf{32.1} & \bf{92.4} & {65.8}   \\
 \hline
\end{tabular}
\end{adjustbox}
\end{table*}

\begin{table*}[t]
\caption{Results of our proposed method and state-of-the-art LiDAR Segmentation methods on SemanticKITTI multi-scan test set - Part \uppercase\expandafter{\romannumeral2}. Note that ``mov car" indicates moving car, and so on.}
\label{semantickitti-multiscan2}
\centering
\begin{adjustbox}{width=\textwidth}
\begin{tabular}{c|c|c|c|c|c|c|c|c|c|c|c|c}
\hline
\textbf{Methods} & \textbf{mIoU} & \rotatebox{90}{vegetation} & \rotatebox{90}{trunk} & \rotatebox{90}{terrain} & \rotatebox{90}{pole} & \rotatebox{90}{traffic} & \rotatebox{90}{mov car} &  \rotatebox{90}{mov truck} & \rotatebox{90}{moving other} & \rotatebox{90}{mov person} & \rotatebox{90}{mov biclist} & \rotatebox{90}{mov motorlist}\\
\hline
\hline
TangentConv~\cite{tatarchenko2018tangent} & 34.1 & 79.5 & 43.2 & 56.7 & 36.4 & 31.2 & 40.3 & 1.1 & 6.4 & 1.9 & 30.1 & 42.2 \\ 
 \hline
DarkNet53~\cite{semantickitti} & 41.6 & 78.4 & 50.7 & 64.8 & 38.1 & 53.3 & 61.5 & 14.1 & 15.2 & 0.2 & 28.9 & 37.8 \\ 
 \hline
SpSeqnet~\cite{shi2020spsequencenet} & 43.1 & 84.0 & 66.0 & 65.7 & 50.8 & 48.7 & 53.2 & \bf{41.2} & \bf{26.2} & 36.2 & 2.3 & 0.1 \\ 
\hline
KPConv~\cite{thomas2019kpconv} &51.2& 84.6& 70.3 & 66.0 & 57.0 & 53.9 & 69.4 & 0.5 & 0.5 & \bf{67.5} & {67.4}&  \bf{47.2} \\
\hline
\hline
Ours & \bf{51.5} & \bf{85.4} & \bf{72.8} & \bf{68.1} & \textbf{62.6} & \bf{61.3} & \bf{68.1} & {0.0} & {0.1} & {63.1} & \bf{60.0} & {0.4}  \\
 \hline
\end{tabular}
\end{adjustbox}
\end{table*}

\myparagraph{Results on SemanticKITTI Multi-scan Semantic Segmentation}
Unlike the single-scan semantic segmentation, the multi-scan segmentation in SemanticKITTI takes multiple frame point cloud as input and generates the more categories under moving status, including moving car, moving truck, moving other-vehicle, moving person, moving bicyclist and moving motorcyclist. In this experiment, we first perform the multiple-frame point cloud fusion. Specifically, the sequential point clouds in LiDAR coordinate are firstly transformed to global coordinate. Then, these sequential point clouds are fused in the global coordinate. Finally, all these points are transformed to the coordinate of last frame. In this way, we can achieve the multiple-frame fusion and we use 3 sequential point clouds as input data in our implementation. We show an example in Fig.~\ref{fig:multiscan}. It can be found that moving cars have multiple shifting point clouds while stationary cars keep all points in same location. 

The results of multi-scan semantic segmentation are shown in Table~\ref{semantickitti-multiscan} and \ref{semantickitti-multiscan2}. Generally, our method outperforms all existing methods in terms of mIoU, where it achieves 0.3\% and 8.4\% gain compared to KPConv~\cite{thomas2019kpconv} (ICCV2019) and SpSeqnet~\cite{shi2020spsequencenet} (CVPR2020), respectively. Our method obtains superior performance for most categories, even for some small objects, like bicycle and motorcycle, \etc.
For these moving categories, our method achieves the best performance on moving car and moving truck.

\myparagraph{Results on nuScenes}
For nuScenes LiDARseg dataset, we report the results on its validation set. As shown in Table~\ref{nuscenes}, our method achieves better performance than existing methods in all categories, and this consistent performance improvement demonstrates the capability of the proposed model. Specifically, the proposed method obtains about 4\% $\thicksim$ 7\% performance gain than projection-based methods. Moreover, for these categories with sparse points, such as bicycle and pedestrian, our method significantly outperforms existing approaches, which also demonstrates the effectiveness of the proposed method to tackle the sparsity and varying density. 
Note that RangeNet++~\cite{milioto2019rangenet++} and Salsanext~\cite{cortinhal2020salsanext} perform the post-processing, including KNN, \etc.

\begin{table*}[t]
\caption{Results of our proposed method and other methods on A2D2 dataset - Part \uppercase\expandafter{\romannumeral1}.}
\label{a2d21}
\centering
\begin{adjustbox}{width=\textwidth}
\begin{tabular}{c|c|c|c|c|c|c|c|c|c|c|c|c|c|c|c|c|c|c}
\hline
\textbf{Methods} & \textbf{mIoU} & \rotatebox{90}{car} &  \rotatebox{90}{bicycle} & \rotatebox{90}{pedestrian} & \rotatebox{90}{truck} & \rotatebox{90}{small-vehi} & \rotatebox{90}{traffic-signal}& \rotatebox{90}{traffic-sign} & \rotatebox{90}{utility-vehi} & \rotatebox{90}{sidebars} & \rotatebox{90}{bumper} & \rotatebox{90}{curbstone} & \rotatebox{90}{solid line} & \rotatebox{90}{irrelevant signs} &
\rotatebox{90}{road blocks} & \rotatebox{90}{tractor} & \rotatebox{90}{non-drivable} & \rotatebox{90}{zebra crossing} \\
\hline
\hline
Squeezeseg~\cite{wu2018squeezeseg} & 8.9 & 9.7 & 0.0 & 0.0 & 15.8 & 0.0 & 0.7 & 64.4 &  0.0 & 0.4 & {0.0} & {2.2} & 15.6 & 0.5 & 15.9 & 0.0 & 0.0 & 0.0 \\
\hline
Squeezesegv2~\cite{wu2019squeezesegv2} & 16.4 & 15.4 & 0.2 & 8.6 & 63.8 & 0.0 & 16.8 & 61.7 & 0.6 & 0.1 & 0.0 & 14.8 & 24.7 & 12.7 & {33.2} & 0.0 & 5.8 & 0.0  \\
\hline
DarkNet53~\cite{semantickitti} & 17.2 & 15.2 & 0.8 & 6.1 & 68.5 & 0.0 & 15.5 & 63.8 & 0.4 & 0.3 & 0.0 & 17.3 & 23.8 & 13.3 & 35.6 & 0.0 & 6.3 & 0.0 \\
\hline
PolarNet~\cite{zhang2020polarnet} & 23.9 & \bf{23.8} & 10.1 & 18.2 & 69.7 & 9.6 & 49.1 & 58.5 & 0.0 & \bf{11.3} & 0.0 & 28.3 & 37.6 & 24.8 & 42.8 & 0.0 & 14.8 & 0.0 \\
\hline
\hline
Ours & \bf{27.1} & 20.1 & \bf{21.1} & \bf{29.6} & \bf{77.8} & \textbf{26.7} & \bf{58.6} & \bf{67.6} & \bf{4.6} & 10.3 & 0.0 & \bf{33.4} & \bf{43.9} & \bf{29.9} & \bf{51.0} & 0.0 & \bf{19.9} & 0.0 \\ 
 \hline
\end{tabular}
\end{adjustbox}
\end{table*}


\begin{table*}[t]
\caption{Results of our proposed method and other methods on A2D2 dataset - Part \uppercase\expandafter{\romannumeral2}.}
\label{a2d2}
\centering
\begin{adjustbox}{width=\textwidth}
\begin{tabular}{c|c|c|c|c|c|c|c|c|c|c|c|c|c|c|c|c|c|c|c|c|c|c}
\hline
\textbf{Methods} & \textbf{mIoU} &   \rotatebox{90}{Obstacles} & \rotatebox{90}{Poles} & \rotatebox{90}{RD restricted area} & \rotatebox{90}{Animals} & \rotatebox{90}{Grid structure} & \rotatebox{90}{Signal corpus} & \rotatebox{90}{Drivable cobbleston} & \rotatebox{90}{Electronic traffic} & \rotatebox{90}{Slow drive area} & \rotatebox{90}{Nature object} & \rotatebox{90}{Parking area} & \rotatebox{90}{Sidewalk} & \rotatebox{90}{Ego car} & \rotatebox{90}{Painted driv instr} & \rotatebox{90}{Traffic guide obj} & \rotatebox{90}{Dashed line} & \rotatebox{90}{RD normal street} & \rotatebox{90}{Sky} & \rotatebox{90}{Buildings} & \rotatebox{90}{Blurred area} & \rotatebox{90}{Rain dirt} \\
\hline
\hline
Squeezeseg~\cite{wu2018squeezeseg} & 8.9 & 0.0& 0.3& 0.0& 0.0& 0.0& 0.0& 0.0& 0.0& 0.0& 64.5& 0.0& 13.7& 0.0& 0.0& 0.1& 0.2& 77.7& 10.4& 27.7& 0.0& 0.0 \\
\hline
Squeezesegv2~\cite{wu2019squeezesegv2} & 16.4 & 0.2& 5.2& 29.5& 0.0& 10.3& 5.5& 2.7& 0.0& 1.9& 76.4& 3.8& 29.2& 0.0& 6.4& 12.4& 17.1& 85.8& 12.1& 50.9& 0.0& 0.0  \\
\hline
DarkNet53~\cite{semantickitti} & 17.2 & 3.9& 7.6& 38.7& 0.0& 10.8& 4.4& 3.3& 0.0& 0.0& 77.9& 3.1& 31.5& 0.0& 9.4& 7.3& 15.7& 86.4& \bf{12.9} & 55.2& 0.0& 0.0 \\
\hline
PolarNet~\cite{zhang2020polarnet} & 23.9 & 8.0& 11.0& 55.6& 0.0& 14.8& 11.9& 7.0& 0.0& \bf{4.4} & 81.6& \bf{12.8} & 42.5& 0.0& 12.7& 11.5& 31.8& 90.3& 9.2& 57.0& 0.0& 0.0 \\
\hline
\hline
Ours & \bf{27.1} & \bf{15.6} & \bf{14.0} & \bf{60.8} & 0.0 & \bf{23.4} & \bf{21.1} & \bf{13.5} & \bf{0.41} & 0.29 & \bf{84.8} & 10.2 & \bf{46.7} & 0.0 & \bf{20.0} & \bf{23.8} & \bf{34.6} & \bf{91.3} & 10.0 & \bf{64.0} & 0.0 & 0.0  \\
 \hline
\end{tabular}
\end{adjustbox}
\end{table*}

\myparagraph{Results on A2D2}
We report the results on A2D2~\cite{geyer2019a2d2} validation set. As shown in Table~\ref{a2d21} and~\ref{a2d2}, it can be observed that the proposed method performs much better than existing methods about 3\% in terms of mIoU, including Squeezeseg~\cite{wu2018squeezeseg}, SqueezesegV2~\cite{wu2019squeezesegv2}, DarkNet53~\cite{semantickitti} and PolarNet~\cite{zhang2020polarnet}, where all of them are based on the 2D projection and 2D convolution networks. Specifically, our method achieves better performance on almost all categories consistently, which also demonstrates the effectiveness of our method. Note that due to the more fine-grained categories in A2D2 (38 categories in total), it is harder than other datasets, such as SemanticKITTI and nuScenes, and there exist more categories with zero values.

In general, our method achieves the consistent state-of-the-art performance in all three datasets with different settings (single-scan and multi-scan) and sensor ranges. It clearly demonstrates the effectiveness of the proposed method and its good generalization capability across different datasets.

\subsection{LiDAR-based Panoptic Segmentation}
Panoptic segmentation is first proposed in \cite{kirillov2019panoptic} as a new task, in which semantic segmentation is performed for background classes and instance segmentation for foreground classes and these two groups of category are also termed as {\bf{stuff}} and {\bf{things}} classes, respectively.
Behley \etal \cite{behley2020benchmark} extend the task to LiDAR point clouds and propose the LiDAR panoptic segmentation. In this experiment, we conduct the panoptic segmentation on SemanticKITTI dataset and report results on the validation set. For the evaluation metrics, we follow the metrics defined in \cite{behley2020benchmark}, where they are the same as that of image panoptic segmentation defined in \cite{kirillov2019panoptic} including Panoptic Quality (PQ), Segmentation Quality (SQ) and Recognition Quality (RQ) which are calculated across all classes. \PQda{} is defined by swapping \PQ{} of each {\bf{stuff}} class to its IoU and averaging over all classes like \PQ{} does. Since the categories in panoptic segmentation contain two groups, \ie, {\bf{stuff}} and {\bf{things}}, these metrics are also performed separately on these two groups, including PQ\textsuperscript{Th}, PQ\textsuperscript{St}, RQ\textsuperscript{Th}, RQ\textsuperscript{St}, SQ\textsuperscript{Th} and SQ\textsuperscript{St}, where Panoptic Quality (PQ) is usually used as the first criteria. For the experimental setting, we follow the LiDAR semantic segmentation, where Adam optimizer with learning rate = $1e^{-4}$ is used for optimization.

In this experiment, we use the proposed cylindrical partition as the partition method and asymmetrical 3D convolution networks as the backbone. Moreover, a semantic branch is used to output the semantic labels for stuff categories, and an instance branch is introduced to generate the instance-level features and further extract their instance IDs for things categories through heuristic clustering algorithms (we use mean-shift in the implementation and the bandwidth of the Mean Shift used in our backbone method is set to $1.2$ while the minimum number of points in a valid instance is set to 50 for SemanticKITTI).

\begin{figure*}[t]
    \centering
    \includegraphics[width=0.9\linewidth]{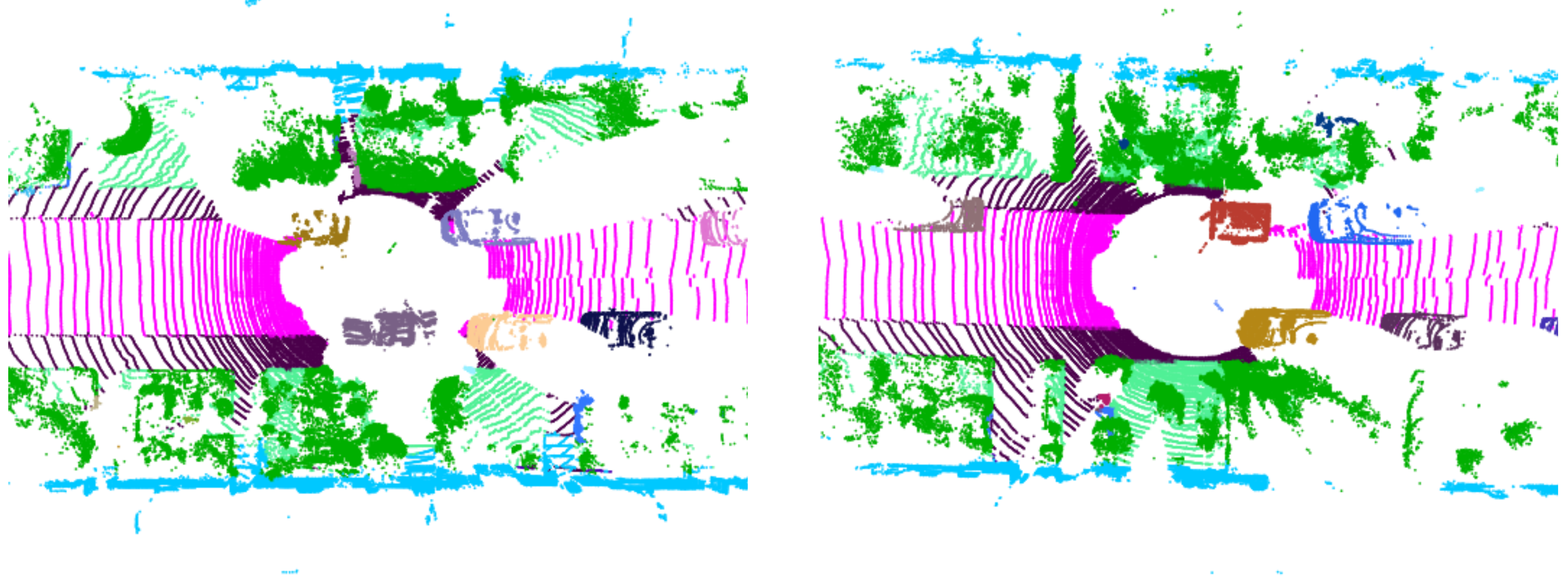}
    \vspace{-2ex}
    \caption{LiDAR-based panoptic segmentation results on SemanticKITTI dataset. For different vehicles, we use different colors to distinguish them.}
    \label{fig:pano}
\end{figure*}

We report the results in Table~\ref{tab:semkitti_val}. It can be found that our method achieves much better performance than existing methods~\cite{milioto2020iros,jiang2020pointgroup}. In terms of PQ, we have about 4.7\% point improvement, and particularly for the thing categories, our method significantly outperforms state-of-the-art in terms of PQ\textsuperscript{Th} and RQ\textsuperscript{Th} with a large margin of 10\% points. It indicates that our cylindrical partition and asymmetrical 3D convolution networks significantly benefit the recognition of the {\bf{things}} classes.
It is worthy of noting that PointGroup and LPASD perform poorly on the outdoor LiDAR segmentation task which indicates that these indoor methods are not suitable for the challenging outdoor point clouds due to the different scenarios and inherent properties.
Experimental results demonstrate the effectiveness of the proposed method and its good generalization ability. We show several samples of panoptic segmentation results in Fig.~\ref{fig:pano}, where different colors represent different vehicles.

\begin{table*}[ht]
\caption{LiDAR-based panoptic segmentation results on the validation set of SemanticKITTI.}
\vspace{-3ex}
    \begin{center}
    \small{
        \begin{tabular}{l|c|ccc|ccc|ccc|c}
            \hline
            Method & \textbf{\PQ} & \PQda & \RQ & \SQ & \PQth & \RQth & \SQth & \PQst & \RQst & \SQst & \miou \\
            \hline\hline
            KPConv \cite{thomas2019kpconv} +
            PV-RCNN \cite{shi2020pv}              & 51.7         & 57.4          & 63.1          & \textbf{78.9} & 46.8          & 56.8          & \textbf{81.5} & \textbf{55.2} & \textbf{67.8} & \textbf{77.1} & 63.1          \\
            PointGroup \cite{jiang2020pointgroup} & 46.1         & 54.0          & 56.6          & 74.6          & 47.7          & 55.9          & 73.8          & 45.0          & 57.1          & 75.1                      & 55.7          \\
            LPASD \cite{milioto2020iros}             & 36.5 & 46.1 & -    & -    & -    & 28.2 & -    & -    & -    & -    & 50.7 \\
            \hline
            \hline
            Ours                          & \textbf{56.4} & \textbf{62.0} & \textbf{67.1} & 76.5          & \textbf{58.8} & \textbf{66.8} & 75.8          & 54.8          & 67.4          & \textbf{77.1} & \textbf{63.5} \\
            \hline
        \end{tabular}
    }
    \end{center}
    \label{tab:semkitti_val}
\end{table*}

\begin{table}[t]
    \caption{LiDAR-based 3D detection results on nuScenes dataset. CyAs denotes the {\bf{Cy}}lindrical partition and {\bf{As}}ymmetrical 3D convolution networks. SSNv2\textcolor{red}{$^\star$} is extracted from nuScenes Leaderboard.}
            \centering
            \begin{tabular*}{0.8\linewidth}{l|c|c}
            \hline
            Methods  & mAP & NDS \\
            \hline
            \hline
            PointPillar~\cite{lang2019pointpillars} & 30.5 & 45.3 \\
            \hline
            PP + Reconfig~\cite{wang2020reconfigurable} & 32.5 & 50.6 \\
            \hline
            SECOND~\cite{yan2018second} & 31.6 & 46.8\\
            \hline
            SECOND~\cite{yan2018second} + Cylinder & 34.3 & 49.6\\
            \hline
            SECOND~\cite{yan2018second} + Asym-CNN & 33.0 & 48.3\\
            \hline
            SECOND~\cite{yan2018second} + CyAs & 36.4 & 51.7\\
            \hline
            SSN~\cite{zhu2020ssn}&  46.3 & 56.9 \\
            \hline
            SSN~\cite{zhu2020ssn} + CyAs & 47.7 & 58.2 \\
            \hline
            SSNv2~\cite{zhu2020ssn}\textcolor{red}{$^\star$} &  50.6 & 61.6 \\
            \hline
            SSNv2~\cite{zhu2020ssn}\textcolor{red}{$^\star$} + CyAs & 52.8 & 64.0 \\
            \hline
            \end{tabular*}
    \label{tab:gene_nusc}
\end{table}

\begin{figure*}[t]
    \centering
    \includegraphics[width=0.9\linewidth]{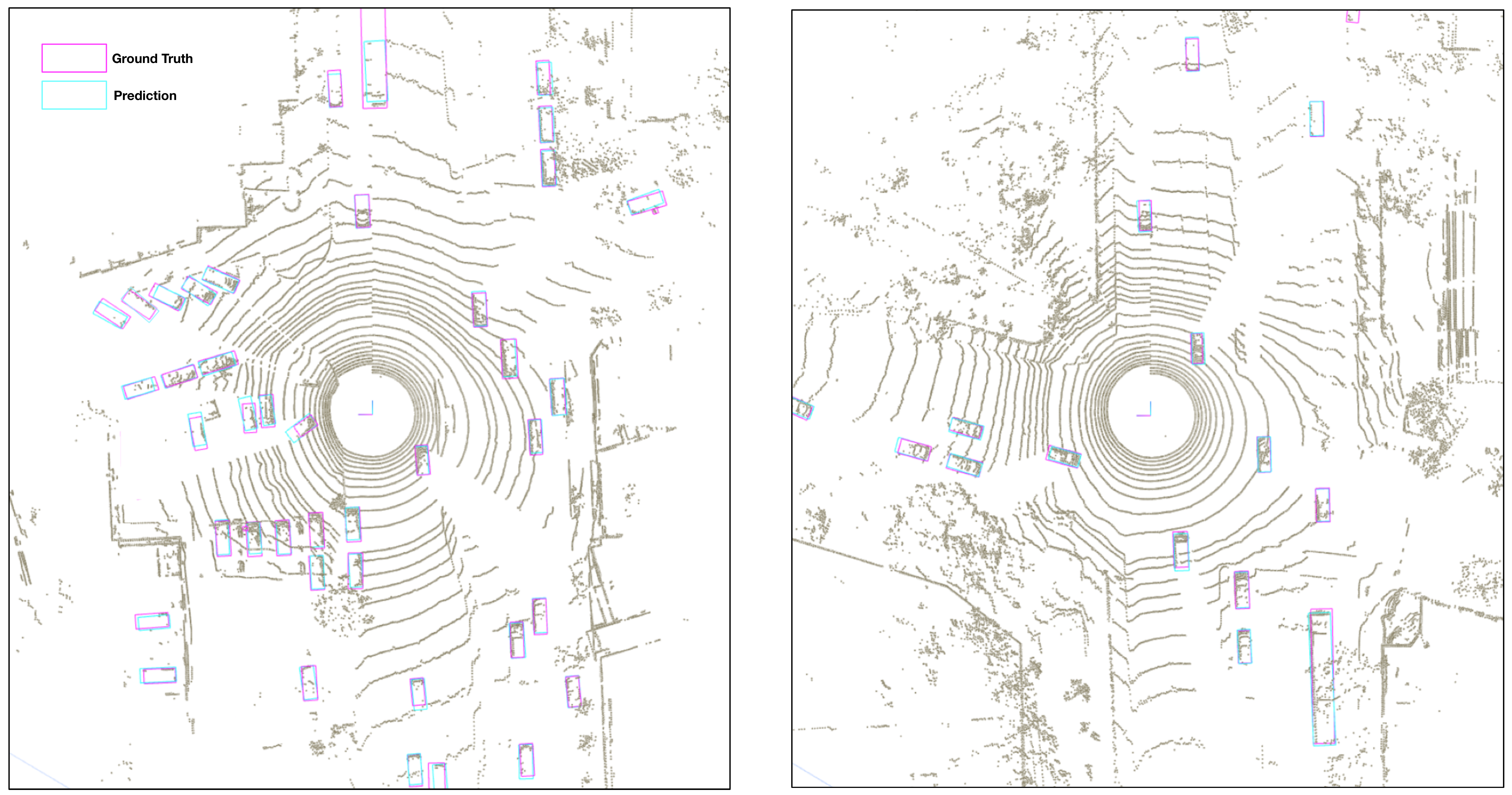}
    \caption{LiDAR-based 3D detection results on nuScenes dataset. Bounding boxes with Pink color and Blue color denote the ground truth and predictions, respectively.}
    \label{fig:3ddet}
\end{figure*}

\subsection{LiDAR-based 3D Detection}
 LiDAR 3D detection aims to localize and classify the multi-class objects in the point cloud. SECOND~\cite{yan2018second} first utilizes the 3D voxelization and 3D convolution networks to perform the single-stage 3D detection. In this experiment, we follow SECOND method and replace the regular voxelization and 3D convolution with the proposed cylindrical partition and asymmetrical 3D convolution networks, respectively. Similarly, to verify its scalability, we also extend the proposed modules to SSN~\cite{zhu2020ssn}. Furthermore, another strong baseline, SSNv2~\cite{zhu2020ssn}, is also adapted to verify the effectiveness of our method when the baseline is very competitive.
The experiments are conducted on nuScenes dataset and the cylindrical partition also generates the $480\times360\times32$ representation. For the evaluation metrics, we follow the official metrics defined in nuScenes, \ie, mean average precision (mAP) and nuScenes detection score (NDS). For other experimental settings, including the optimization method, target assignment, anchor size and network architecture of multiple heads, we all follow the setting in SSN~\cite{zhu2020ssn}.

The results are shown in Table~\ref{tab:gene_nusc}. PP + Reconfig~\cite{wang2020reconfigurable} is a partition enhancement approach based on PointPillar~\cite{lang2019pointpillars}, while our SECOND + CyAs performs better with similar backbone, which indicates the superiority of the cylindrical partition. To verify the effect of different components (\ie, Cylinder partition and Asymmetrical 3D convolution networks) of our method on LiDAR 3D detection, we design two variants, \ie, SECOND~\cite{yan2018second} + Cylinder and SECOND~\cite{yan2018second} + Asym-CNN. The results shown in Table~\ref{tab:gene_nusc} demonstrate that these two components in our method consistently improve the baseline method with 2.8\% points and 1.5\% points in terms of NDS, respectively.
 We then extend the proposed method (\ie, CyAs) to two baseline methods, termed as SECOND + CyAs and SSN + CyAs, respectively.
 By comparing these two models with their extensions, it can be observed that the proposed Cylindrical partition and Asymmetrical 3D convolution networks boost the performance consistently, even for the strong baseline \ie, SSNv2, which demonstrates the effectiveness and scalability of our model. For different backbones, like SECOND and SSN, our method could consistently benefit them, showing its good generalization ability. Several qualitative results on nuScenes dataset are shown in Fig.~\ref{fig:3ddet}.

\subsection{Ablation Studies}

In this section, we perform the thorough ablation experiments on LiDAR-based semantic segmentation task to investigate the effect of different components in our method. We also design several variants of asymmetrical residual block to verify our claim that strengthening the horizontal and vertical kernels power the representation ability for driving-scene point cloud. For the 3D representation $480\times360\times32$ after cylindrical partition, we also try several other hyper-parameters to cross-validate these values.

\begin{table}[t]
\caption{Ablation studies for network components on SemanticKITTI validation set. PR denotes the point-wise refinement module. Asym-CNN denotes the asymmetrical residual block.}
\label{table_net_components}
\centering
\setlength{\tabcolsep}{4.2pt}
\begin{tabular*}{1.0\linewidth}{c c c c c | c}
\hline
{Baseline} & Cylinder & {Asym-CNN} & {DDCM} & {PR} & {mIoU} \\
\hline
\hline
{\cmark} & & & & & 58.1 \\
\cmark & \cmark & & & & 61.0 \\
\cmark & \cmark & \cmark & & & 63.8\\
\cmark & \cmark & \cmark & \cmark & & 65.2 \\
\cmark & \cmark & \cmark & \cmark & \cmark & 65.9 \\
\hline
\end{tabular*}
\end{table}

\myparagraph{Effects of Network Components}
In this part, we make several variants of our model to validate the contributions of different components. The results on SemanticKITTI validation set are reported in Table~\ref{table_net_components}. Baseline method denotes the framework using 3D voxel partition (with cubic partition) and 3D convolution networks. It can be observed that cylindrical partition performs much better than cubic-based partition with about 3\% mIoU gain and asymmetrical 3D convolution networks also significantly boost the performance about 3\% improvement, which demonstrates that both cylindrical partition and asymmetrical 3D convolution networks are crucial in the proposed method. Furthermore, dimension-decomposition based context modeling delivers the effective global context features, which yields an improvement of 1.4\%. Point-wise refinement module further pushes forward the performance based on the strong model, about 0.7\%. Generally, the proposed cylindrical partition and asymmetrical 3D convolution networks make the most contribution to the performance improvement.

\begin{table*}[t]
\caption{Ablation studies for asymmetrical residual block on SemanticKITTI validation set.}
\label{tab:asym_vari}
\centering
\begin{adjustbox}{width=\textwidth}
\begin{tabular}{c|c|c|c|c|c|c|c|c|c|c|c|c|c|c|c|c|c|c|c|c}
\hline
\textbf{Methods} & \textbf{mIoU} & \rotatebox{90}{car} &  \rotatebox{90}{bicycle} & \rotatebox{90}{motorcycle} & \rotatebox{90}{truck} & \rotatebox{90}{other-vehicle} & \rotatebox{90}{person} & \rotatebox{90}{bicyclist} & \rotatebox{90}{motorcyclist} & \rotatebox{90}{road} & \rotatebox{90}{parking} & \rotatebox{90}{sidewalk} & \rotatebox{90}{other-ground} &
\rotatebox{90}{building} & \rotatebox{90}{fence} & \rotatebox{90}{vegetation} & \rotatebox{90}{trunk} & \rotatebox{90}{terrain} & \rotatebox{90}{pole} & \rotatebox{90}{traffic} \\
\hline
\hline
Regular & 60.8 & 96.7 & 43.5 & 50.6 & 78.0 & 56.4 & 64.5 & 81.7 & 0.1 & 93.5 & 38.2 & 78.6 & 0.2 & 89.5 & 54.0 & 86.9 & 61.4 & 71.8 & 62.8 & 46.2 \\
\hline
1D-asymmetrical & 61.9 & 96.7 & 44.3 & 56.4 & 83.2 & 60.6 & 64.8 & 89.9 & 2.3 & 93.1 & 36.2 & 78.0 & 0.4 & 88.5 & 49.5 & 86.7 & 64.8 & 70.6 & 63.3 & 48.7 \\
\hline
\hline
Asymmetrical(ours) & 63.9 & 96.8 & 54.4 & 76.1 & 80.1 & 65.3 & 74.2 & 88.7 & 6.2 & 90.1 & 36.6 & 75.1 & 0.4 & 83.3 & 55.2 & 86.4 & 67.3 & 67.1 & 63.4 & 46.7  \\
 \hline
\end{tabular}
\end{adjustbox}
\end{table*}

\begin{figure}[t]
    \centering
    \includegraphics[width=1.0\linewidth]{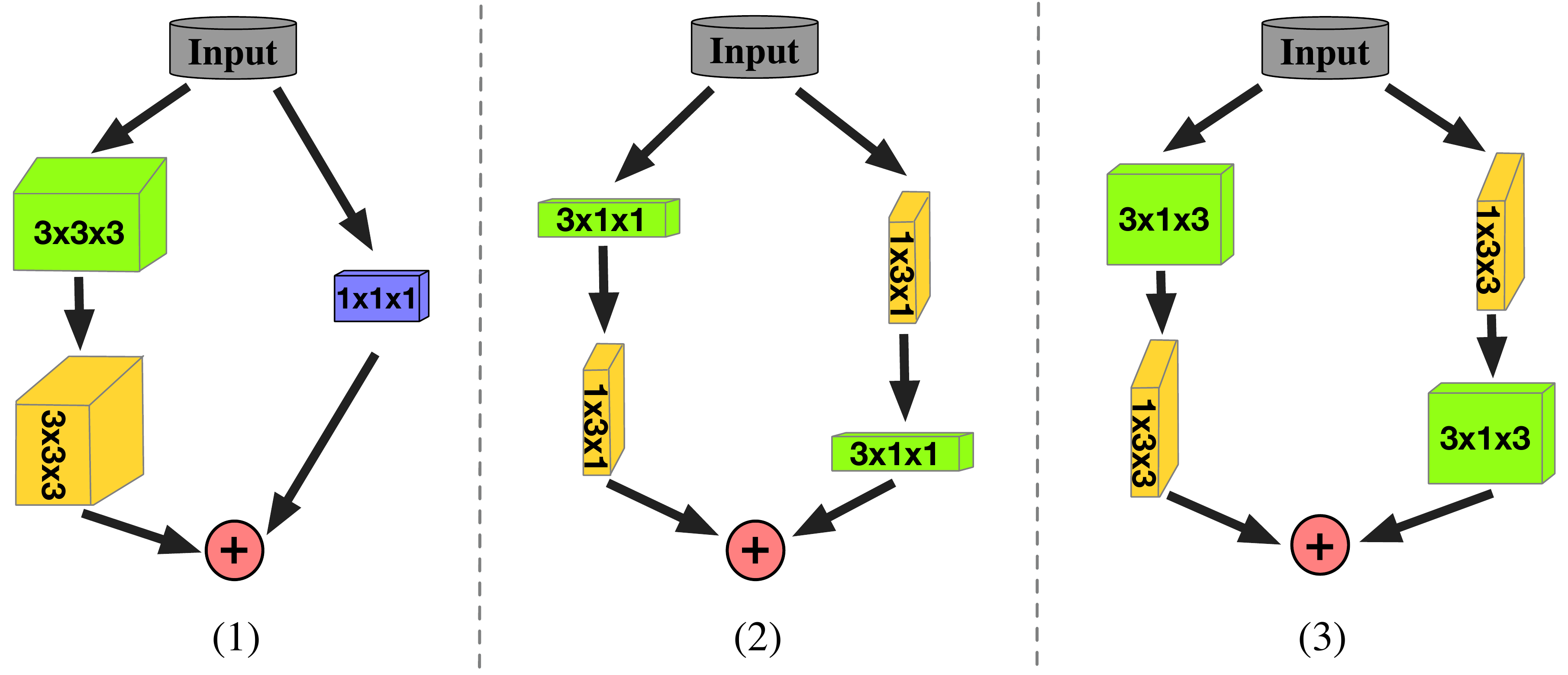}
    \caption{We design three blocks for ablation studies of asymmetrical residual block, including (1) regular residual block , (2) 1D-asymmetrical residual block without height and (3) the proposed asymmetrical residual block.}
    \label{fig:asym_vari}
\end{figure}

\myparagraph{Variants of Asymmetrical Residual Block}
To verify the effectiveness of the proposed asymmetrical residual block, we design several variants of asymmetrical residual block to investigate the effect of horizontal and vertical kernel enhancement (as shown in Fig.~\ref{fig:asym_vari}). The first variant is the regular residual block without any asymmetrical structure. The second one is the 1D-asymmetrical residual block, which utilizes the 1D asymmetrical kernels without height and also strengthens the horizontal or vertical kernels in one-dimension. The third one is the proposed asymmetrical residual block, which strengthens both horizontal and vertical kernels. These variants strengthen the skeleton of convolution kernels step by step (from regular residual block to asymmetrical kernel without height, then to both horizontal and vertical kernels with height).

We conduct the ablation studies on SemanticKITTI validation set. Note that we use the cylindrical partition as the partition method and stack these proposed variants to build the 3D convolution networks for this ablation experiment. We report the results in Table~\ref{tab:asym_vari}. It can be found that although the 1D-Asymmetrical residual block only powers the horizontal and vertical kernels in one-dimension, it still achieves 1.3\% gain in terms of mIoU and it obtains about more than 5\% performance gain for motorcycle, other-vehicle and bicyclist, which demonstrates the effectiveness of strengthening skeleton of convolution kernel, even without height dimension. After taking the height into the consideration, the proposed asymmetrical residual block further matches the object distribution in driving scene and powers the skeleton of kernels, which enhances the robustness to the sparsity. From Table~\ref{tab:asym_vari}, the proposed asymmetrical residual block significantly boosts the performance with about 3\% improvements, where large improvement can be observed on some instance categories (about 10\% gain), including bicycle, person, other-vehicle and motorcycle, because it matches the point distribution of object and enhances the representational power.

\myparagraph{Size of 3D Representation}
As mentioned in implementation details, we set the size of 3D representation to $480\times360\times32$. In this experiment, we use other hyper-parameters to cross-validate these values, including $600\times480\times40$ and $320\times240\times24$. They cover the denser and sparser representations compared with $480\times360\times32$ representation. Furthermore, we also introduce a cubic partition with size of $1000\times1000\times45$ as the counterpart to investigate the effectiveness and compactness of cylindrical partition.

We conduct the experiments on SemanticKITTI validation set and all experiments are under same settings except the different representation size. The results are shown in Table~\ref{tab:sizeofrep}. It can be found that the 3D representation with $480\times360\times32$ performs better than other two representations with 2\% point improvement than $600\times480\times40$ and $320\times240\times24$. Since $320\times240\times24$ representation delivers compacter representation with larger cylindrical cells, it however might mis-split the points across different categories into same cell, which 
inevitably increases the information loss; While for $600\times480\times40$ representation, it contains fine-grained cylindrical cell, but generates the larger representation, which might burden the training of 3D convolution and cause the degradation of performance. Compared to the cubic partition with $1000\times1000\times45$, all cylindrical partitions achieve much better performance and this consistent performance gain demonstrates its effectiveness.
From this experiment, we cross-validate the $480\times360\times32$ representation and investigate the effect of different size of 3D representations. 

\begin{table}
\caption{Ablation studies for the size of 3D representation on SemanticKITTI validation set.}
\setlength{\tabcolsep}{7.0pt}
\centering
\begin{tabular*}{0.7\linewidth}{c|c}
\hline
Size of Representation  & mIoU \\
\hline
\hline
 $1000\times1000\times45$ (cubic) & 58.1\\
 \hline
 $320\times240\times24$ & 63.4\\
\hline
$600\times480\times40$ & 64.1 \\
\hline
$480\times360\times32$ & 65.9 \\
\hline
\end{tabular*}
\label{tab:sizeofrep}
\end{table}

\section{Discussion}


\subsection{Comparison of Inference Time}

\begin{table}[t]
\vspace{-0.5ex}
\caption{Comparison of inference time on SemanticKITTI dataset. All numbers represent the total time consisting of computation time and post-processing time. Results are directly taken from~\cite{semantickitti}.}
\centering
\begin{tabular*}{0.82\linewidth}{l|c|c}
\hline
Methods  & Latency (ms) & mIoU(\%) \\
\hline
\hline
TangentConv~\cite{tatarchenko2018tangent} & 3000 & 40.9 \\
\hline
RandLA~\cite{hu2020randla} & 800 & 53.9 \\
\hline
KPConv~\cite{thomas2019kpconv} & 263 & 58.8\\
\hline
MinkowskiNet~\cite{choy20194d} & 294 & 63.1 \\
\hline
SPVNAS-lite~\cite{tang2020searching} & 110 & 63.7\\
\hline
SPVNAS~\cite{tang2020searching} & 259 & 66.4\\
\hline
\hline
Ours & 170 & 67.8\\
\hline
\end{tabular*}
\label{tab:time1}
\end{table}

To investigate the efficiency of the proposed method, we further make a statistic of inference time compared to existing methods. In the experiment, we keep the setting unchanged and set the mode as the evaluation mode to calculate the inference time. The results of inference time of existing methods are directly token from~\cite{semantickitti}.

The results are shown in Table~\ref{tab:time1}. Compared to 2D projection based method (inference time consists of computation time and post-processing time), \ie, RandLA~\cite{hu2020randla}, our method achieves about \textbf{5.0}$\times$ speedup with 14\% performance improvement due to no requirement for post-processing. 
Moreover, compared to other 3D based methods, including MinkowskiNet~\cite{choy20194d} and SPVNAS~\cite{tang2020searching}, we also achieve the better performance and less inference time. The main reasons lie in two aspects: 1) the proposed cylindrical partition generates compacter representation compared to regular cubic partition. For example, the regular cubic partition often has the cell of $0.1 \times 0.1 \times 0.1$, and it thus generates a 3D representation of $1000\times1000\times45$, which is more than 4 times larger than the cylindrical partition. 2) the asymmetrical 3D convolution networks consume smaller computational overhead and less parameters compared to the regular 3D convolution networks. Specifically, using a convolution with kernel=$3\times1\times3$ followed by a $1\times3\times3$ convolution is equivalent to sliding a two layer network with the same receptive field as in a 3D convolution with kernel= $3\times3\times3$, but it has 33\% cheaper computational cost than a $3\times3\times3$ convolution with same number of output filters. The corporation of these two parts leads to the effective and efficient approach.

\subsection{Comparison with other methods dealing with sparsity issue}

Our proposed method utilizes the cylindrical partition and asymmetrical 3D convolution networks to handle the inherent difficulties, \ie, sparsity and varying density. Hence, we further compare the proposed method with other methods tackling the sparsity issue, to verify its effectiveness. Specifically, SPVNAS~\cite{tang2020searching} proposes a sparse point-voxel convolution to preserve the fine details and deal with sparsity. MinkowskiNet~\cite{choy20194d} adopts sparse tensors and proposes a generalized sparse convolution. We take them as the counterpart dealing with the sparsity issue and make a comparison with them.
Note that in our implementation, we also use the sparse convolution~\cite{yan2018second} to build up the asymmetrical 3D convolution networks.

\begin{table}[t]
\vspace{-0.5ex}
\caption{Comparison with other methods dealing with sparsity issue on SemanticKITTI validation set.}
\setlength{\tabcolsep}{3.0pt}
\centering
\begin{tabular*}{1.0\linewidth}{l|c|c|c}
\hline
Methods  &SparseConv& Latency (ms) & mIoU(\%) \\
\hline
\hline
MinkowskiNet~\cite{choy20194d} & \cmark & 294 & 63.1 \\
\hline
SPVNAS~\cite{tang2020searching} & \cmark & 259 & 66.4\\
\hline
\hline
Ours& \cmark & 170 & 67.8\\
\hline
\end{tabular*}
\label{tab:sparse1}
\end{table}

The results are shown in Table~\ref{tab:sparse1}. Compared to other methods handling the sparsity issue, our method achieves both better performance and efficiency, which also demonstrates the superiority of our method.

\section{Conclusion}

In this paper, we have proposed a cylindrical and asymmetrical 3D convolution networks for LiDAR segmentation, where it maintains the 3D geometric relation. Specifically, two key components, the cylinder partition and asymmetrical 3D convolution networks, are designed to handle the inherent difficulties in outdoor LiDAR point cloud, namely sparsity and varying density, effectively and robustly.
We conduct the extensive experiments and ablation studies, where the model achieves the state-of-the-art in SemanticKITTI, A2D2 and nuScenes, and keeps good generalization ability to other LiDAR based tasks, including LiDAR panoptic segmentation and LiDAR 3D detection.


%

%

%
%

\ifCLASSOPTIONcaptionsoff
  \newpage
\fi



%
%
%

%

\bibliographystyle{IEEEtran}
\bibliography{ref}

\begin{thebibliography}{10}
\providecommand{\url}[1]{#1}
\csname url@samestyle\endcsname
\providecommand{\newblock}{\relax}
\providecommand{\bibinfo}[2]{#2}
\providecommand{\BIBentrySTDinterwordspacing}{\spaceskip=0pt\relax}
\providecommand{\BIBentryALTinterwordstretchfactor}{4}
\providecommand{\BIBentryALTinterwordspacing}{\spaceskip=\fontdimen2\font plus
\BIBentryALTinterwordstretchfactor\fontdimen3\font minus
  \fontdimen4\font\relax}
\providecommand{\BIBforeignlanguage}[2]{{%
\expandafter\ifx\csname l@#1\endcsname\relax
\typeout{** WARNING: IEEEtran.bst: No hyphenation pattern has been}%
\typeout{** loaded for the language `#1'. Using the pattern for}%
\typeout{** the default language instead.}%
\else
\language=\csname l@#1\endcsname
\fi
#2}}
\providecommand{\BIBdecl}{\relax}
\BIBdecl

\bibitem{ma2019trafficpredict}
Y.~Ma, X.~Zhu, S.~Zhang, R.~Yang, W.~Wang, and D.~Manocha, ``Trafficpredict:
  Trajectory prediction for heterogeneous traffic-agents,'' in
  \emph{Proceedings of the AAAI Conference on Artificial Intelligence},
  vol.~33, no.~01, 2019, pp. 6120--6127.

\bibitem{xu2019depth}
Y.~Xu, X.~Zhu, J.~Shi, G.~Zhang, H.~Bao, and H.~Li, ``Depth completion from
  sparse lidar data with depth-normal constraints,'' in \emph{Proceedings of
  the IEEE/CVF International Conference on Computer Vision}, 2019, pp.
  2811--2820.

\bibitem{wang2021probabilistic}
T.~Wang, X.~Zhu, J.~Pang, and D.~Lin, ``Probabilistic and geometric depth:
  Detecting objects in perspective,'' \emph{arXiv preprint arXiv:2107.14160},
  2021.

\bibitem{wang2021fcos3d}
------, ``Fcos3d: Fully convolutional one-stage monocular 3d object
  detection,'' \emph{arXiv preprint arXiv:2104.10956}, 2021.

\bibitem{peng2021side}
X.~Peng, X.~Zhu, T.~Wang, and Y.~Ma, ``Side: Center-based stereo 3d detector
  with structure-aware instance depth estimation,'' \emph{arXiv preprint
  arXiv:2108.09663}, 2021.

\bibitem{zhang2020polarnet}
Y.~Zhang, Z.~Zhou, P.~David, X.~Yue, Z.~Xi, B.~Gong, and H.~Foroosh,
  ``Polarnet: An improved grid representation for online lidar point clouds
  semantic segmentation,'' in \emph{CVPR}, 2020, pp. 9601--9610.

\bibitem{hu2020randla}
Q.~Hu, B.~Yang, L.~Xie, S.~Rosa, Y.~Guo, Z.~Wang, N.~Trigoni, and A.~Markham,
  ``Randla-net: Efficient semantic segmentation of large-scale point clouds,''
  in \emph{CVPR}, 2020, pp. 11\,108--11\,117.

\bibitem{milioto2019rangenet++}
A.~Milioto, I.~Vizzo, J.~Behley, and C.~Stachniss, ``Rangenet++: Fast and
  accurate lidar semantic segmentation,'' in \emph{IROS}.\hskip 1em plus 0.5em
  minus 0.4em\relax IEEE, 2019, pp. 4213--4220.

\bibitem{xu2020squeezesegv3}
C.~Xu, B.~Wu, Z.~Wang, W.~Zhan, P.~Vajda, K.~Keutzer, and M.~Tomizuka,
  ``Squeezesegv3: Spatially-adaptive convolution for efficient point-cloud
  segmentation,'' \emph{arXiv preprint arXiv:2004.01803}, 2020.

\bibitem{wu2018squeezeseg}
B.~Wu, A.~Wan, X.~Yue, and K.~Keutzer, ``Squeezeseg: Convolutional neural nets
  with recurrent crf for real-time road-object segmentation from 3d lidar point
  cloud,'' in \emph{ICRA}.\hskip 1em plus 0.5em minus 0.4em\relax IEEE, 2018,
  pp. 1887--1893.

\bibitem{lang2019pointpillars}
A.~H. Lang, S.~Vora, H.~Caesar, L.~Zhou, J.~Yang, and O.~Beijbom,
  ``Pointpillars: Fast encoders for object detection from point clouds,'' in
  \emph{CVPR}, 2019, pp. 12\,697--12\,705.

\bibitem{graham20183d}
B.~Graham, M.~Engelcke, and L.~Van Der~Maaten, ``3d semantic segmentation with
  submanifold sparse convolutional networks,'' in \emph{CVPR}, 2018, pp.
  9224--9232.

\bibitem{cciccek20163d}
{\"O}.~{\c{C}}i{\c{c}}ek, A.~Abdulkadir, S.~S. Lienkamp, T.~Brox, and
  O.~Ronneberger, ``3d u-net: learning dense volumetric segmentation from
  sparse annotation,'' in \emph{MICCAI}.\hskip 1em plus 0.5em minus 0.4em\relax
  Springer, 2016, pp. 424--432.

\bibitem{behley2019semantickitti}
J.~Behley, M.~Garbade, A.~Milioto, J.~Quenzel, S.~Behnke, C.~Stachniss, and
  J.~Gall, ``Semantickitti: A dataset for semantic scene understanding of lidar
  sequences,'' in \emph{ICCV}, 2019, pp. 9297--9307.

\bibitem{nuscenes}
H.~Caesar, V.~Bankiti, A.~H. Lang, S.~Vora, V.~E. Liong, Q.~Xu, A.~Krishnan,
  Y.~Pan, G.~Baldan, and O.~Beijbom, ``nuscenes: A multimodal dataset for
  autonomous driving,'' \emph{arXiv preprint arXiv:1903.11027}, 2019.

\bibitem{geyer2019a2d2}
J.~Geyer, Y.~Kassahun, M.~Mahmudi, X.~Ricou, and R.~Durgesh, ``A2d2: Aev
  autonomous driving dataset,'' \emph{Note: http://www. a2d2. audi}, vol.~1,
  no.~4, 2019.

\bibitem{qi2017pointnet}
C.~R. Qi, H.~Su, K.~Mo, and L.~J. Guibas, ``Pointnet: Deep learning on point
  sets for 3d classification and segmentation,'' in \emph{CVPR}, 2017, pp.
  652--660.

\bibitem{thomas2019kpconv}
H.~Thomas, C.~R. Qi, J.-E. Deschaud, B.~Marcotegui, F.~Goulette, and L.~J.
  Guibas, ``Kpconv: Flexible and deformable convolution for point clouds,'' in
  \emph{ICCV}, 2019, pp. 6411--6420.

\bibitem{wu2019pointconv}
W.~Wu, Z.~Qi, and L.~Fuxin, ``Pointconv: Deep convolutional networks on 3d
  point clouds,'' in \emph{CVPR}, 2019, pp. 9621--9630.

\bibitem{wang2019dynamic}
Y.~Wang, Y.~Sun, Z.~Liu, S.~E. Sarma, M.~M. Bronstein, and J.~M. Solomon,
  ``Dynamic graph cnn for learning on point clouds,'' \emph{Acm Transactions On
  Graphics (tog)}, vol.~38, no.~5, pp. 1--12, 2019.

\bibitem{velivckovic2017graph}
P.~Veli{\v{c}}kovi{\'c}, G.~Cucurull, A.~Casanova, A.~Romero, P.~Lio, and
  Y.~Bengio, ``Graph attention networks,'' \emph{arXiv preprint
  arXiv:1710.10903}, 2017.

\bibitem{lyu2020learning}
Y.~Lyu, X.~Huang, and Z.~Zhang, ``Learning to segment 3d point clouds in 2d
  image space,'' in \emph{CVPR}, 2020, pp. 12\,255--12\,264.

\bibitem{engelmann20203d}
F.~Engelmann, M.~Bokeloh, A.~Fathi, B.~Leibe, and M.~Nie{\ss}ner, ``3d-mpa:
  Multi-proposal aggregation for 3d semantic instance segmentation,'' in
  \emph{CVPR}, 2020, pp. 9031--9040.

\bibitem{zhang2020fusion}
J.~Zhang, C.~Zhu, L.~Zheng, and K.~Xu, ``Fusion-aware point convolution for
  online semantic 3d scene segmentation,'' in \emph{CVPR}, 2020, pp.
  4534--4543.

\bibitem{yan2020pointasnl}
X.~Yan, C.~Zheng, Z.~Li, S.~Wang, and S.~Cui, ``Pointasnl: Robust point clouds
  processing using nonlocal neural networks with adaptive sampling,'' in
  \emph{CVPR}, 2020, pp. 5589--5598.

\bibitem{wang2019graph}
L.~Wang, Y.~Huang, Y.~Hou, S.~Zhang, and J.~Shan, ``Graph attention convolution
  for point cloud semantic segmentation,'' in \emph{CVPR}, 2019, pp.
  10\,296--10\,305.

\bibitem{pham2019jsis3d}
Q.-H. Pham, T.~Nguyen, B.-S. Hua, G.~Roig, and S.-K. Yeung, ``Jsis3d: joint
  semantic-instance segmentation of 3d point clouds with multi-task pointwise
  networks and multi-value conditional random fields,'' in \emph{CVPR}, 2019,
  pp. 8827--8836.

\bibitem{qi20173d}
X.~Qi, R.~Liao, J.~Jia, S.~Fidler, and R.~Urtasun, ``3d graph neural networks
  for rgbd semantic segmentation,'' in \emph{ICCV}, 2017, pp. 5199--5208.

\bibitem{cortinhal2020salsanext}
T.~Cortinhal, G.~Tzelepis, and E.~E. Aksoy, ``Salsanext: Fast,
  uncertainty-aware semantic segmentation of lidar point clouds for autonomous
  driving,'' 2020.

\bibitem{zhao2021lif}
L.~Zhao, H.~Zhou, X.~Zhu, X.~Song, H.~Li, and W.~Tao, ``Lif-seg: Lidar and
  camera image fusion for 3d lidar semantic segmentation,'' \emph{arXiv
  preprint arXiv:2108.07511}, 2021.

\bibitem{alonso20203d}
I.~Alonso, L.~Riazuelo, L.~Montesano, and A.~C. Murillo, ``3d-mininet: Learning
  a 2d representation from point clouds for fast and efficient 3d lidar
  semantic segmentation,'' \emph{arXiv preprint arXiv:2002.10893}, 2020.

\bibitem{zhang12356deep}
F.~Zhang, J.~Fang, B.~Wah, and P.~Torr, ``Deep fusionnet for point cloud
  semantic segmentation,'' \emph{ECCV}, 2020.

\bibitem{landrieu2018large}
L.~Landrieu and M.~Simonovsky, ``Large-scale point cloud semantic segmentation
  with superpoint graphs,'' in \emph{CVPR}, 2018, pp. 4558--4567.

\bibitem{hong2020lidar}
F.~Hong, H.~Zhou, X.~Zhu, H.~Li, and Z.~Liu, ``Lidar-based panoptic
  segmentation via dynamic shifting network,'' \emph{arXiv preprint
  arXiv:2011.11964}, 2020.

\bibitem{cong2021input}
P.~Cong, X.~Zhu, and Y.~Ma, ``Input-output balanced framework for long-tailed
  lidar semantic segmentation,'' in \emph{2021 IEEE International Conference on
  Multimedia and Expo (ICME)}.\hskip 1em plus 0.5em minus 0.4em\relax IEEE,
  2021, pp. 1--6.

\bibitem{wu2019squeezesegv2}
B.~Wu, X.~Zhou, S.~Zhao, X.~Yue, and K.~Keutzer, ``Squeezesegv2: Improved model
  structure and unsupervised domain adaptation for road-object segmentation
  from a lidar point cloud,'' in \emph{ICRA}.\hskip 1em plus 0.5em minus
  0.4em\relax IEEE, 2019, pp. 4376--4382.

\bibitem{han2020occuseg}
L.~Han, T.~Zheng, L.~Xu, and L.~Fang, ``Occuseg: Occupancy-aware 3d instance
  segmentation,'' in \emph{CVPR}, 2020, pp. 2940--2949.

\bibitem{tchapmi2017segcloud}
L.~Tchapmi, C.~Choy, I.~Armeni, J.~Gwak, and S.~Savarese, ``Segcloud: Semantic
  segmentation of 3d point clouds,'' in \emph{3DV}.\hskip 1em plus 0.5em minus
  0.4em\relax IEEE, 2017, pp. 537--547.

\bibitem{meng2019vv}
H.-Y. Meng, L.~Gao, Y.-K. Lai, and D.~Manocha, ``Vv-net: Voxel vae net with
  group convolutions for point cloud segmentation,'' in \emph{ICCV}, 2019, pp.
  8500--8508.

\bibitem{cylinder3d1}
H.~Zhou, X.~Zhu, X.~Song, Y.~Ma, Z.~Wang, H.~Li, and D.~Lin, ``Cylinder3d: An
  effective 3d framework for driving-scene lidar semantic segmentation,''
  \emph{arXiv preprint arXiv:2008.01550}, 2020.

\bibitem{cylinder3d2}
X.~Zhu, H.~Zhou, T.~Wang, F.~Hong, Y.~Ma, W.~Li, H.~Li, and D.~Lin,
  ``Cylindrical and asymmetrical 3d convolution networks for lidar
  segmentation,'' in \emph{Proceedings of the IEEE/CVF Conference on Computer
  Vision and Pattern Recognition}, 2021, pp. 9939--9948.

\bibitem{long2015fully}
J.~Long, E.~Shelhamer, and T.~Darrell, ``Fully convolutional networks for
  semantic segmentation,'' in \emph{CVPR}, 2015, pp. 3431--3440.

\bibitem{chen2017deeplab}
L.-C. Chen, G.~Papandreou, I.~Kokkinos, K.~Murphy, and A.~L. Yuille, ``Deeplab:
  Semantic image segmentation with deep convolutional nets, atrous convolution,
  and fully connected crfs,'' \emph{IEEE transactions on pattern analysis and
  machine intelligence}, vol.~40, no.~4, pp. 834--848, 2017.

\bibitem{chen2018encoder}
L.-C. Chen, Y.~Zhu, G.~Papandreou, F.~Schroff, and H.~Adam, ``Encoder-decoder
  with atrous separable convolution for semantic image segmentation,'' in
  \emph{ECCV}, 2018, pp. 801--818.

\bibitem{zhao2017pyramid}
H.~Zhao, J.~Shi, X.~Qi, X.~Wang, and J.~Jia, ``Pyramid scene parsing network,''
  in \emph{CVPR}, 2017, pp. 2881--2890.

\bibitem{liu2019auto}
C.~Liu, L.-C. Chen, F.~Schroff, H.~Adam, W.~Hua, A.~L. Yuille, and L.~Fei-Fei,
  ``Auto-deeplab: Hierarchical neural architecture search for semantic image
  segmentation,'' in \emph{CVPR}, 2019, pp. 82--92.

\bibitem{tang2020searching}
H.~Tang, Z.~Liu, S.~Zhao, Y.~Lin, J.~Lin, H.~Wang, and S.~Han, ``Searching
  efficient 3d architectures with sparse point-voxel convolution,'' \emph{arXiv
  preprint arXiv:2007.16100}, 2020.

\bibitem{ronneberger2015u}
O.~Ronneberger, P.~Fischer, and T.~Brox, ``U-net: Convolutional networks for
  biomedical image segmentation,'' in \emph{MICCAI}.\hskip 1em plus 0.5em minus
  0.4em\relax Springer, 2015, pp. 234--241.

\bibitem{wang2019shape}
W.~Wang, E.~Xie, X.~Li, W.~Hou, T.~Lu, G.~Yu, and S.~Shao, ``Shape robust text
  detection with progressive scale expansion network,'' in \emph{CVPR}, 2019,
  pp. 9336--9345.

\bibitem{ding2019acnet}
X.~Ding, Y.~Guo, G.~Ding, and J.~Han, ``Acnet: Strengthening the kernel
  skeletons for powerful cnn via asymmetric convolution blocks,'' in
  \emph{ICCV}, 2019, pp. 1911--1920.

\bibitem{zhang2019co}
H.~Zhang, H.~Zhang, C.~Wang, and J.~Xie, ``Co-occurrent features in semantic
  segmentation,'' in \emph{CVPR}, 2019, pp. 548--557.

\bibitem{chen2020tensor}
W.~Chen, X.~Zhu, R.~Sun, J.~He, R.~Li, X.~Shen, and B.~Yu, ``Tensor low-rank
  reconstruction for semantic segmentation,'' \emph{arXiv preprint
  arXiv:2008.00490}, 2020.

\bibitem{berman2018lovasz}
M.~Berman, A.~Rannen~Triki, and M.~B. Blaschko, ``The lov{\'a}sz-softmax loss:
  A tractable surrogate for the optimization of the intersection-over-union
  measure in neural networks,'' in \emph{CVPR}, 2018, pp. 4413--4421.

\bibitem{zhu2020ssn}
X.~Zhu, Y.~Ma, T.~Wang, Y.~Xu, J.~Shi, and D.~Lin, ``Ssn: Shape signature
  networks for multi-class object detection from point clouds,'' \emph{ECCV},
  2020.

\bibitem{yan2018second}
Y.~Yan, Y.~Mao, and B.~Li, ``Second: Sparsely embedded convolutional
  detection,'' \emph{Sensors}, vol.~18, no.~10, p. 3337, 2018.

\bibitem{tatarchenko2018tangent}
M.~Tatarchenko, J.~Park, V.~Koltun, and Q.-Y. Zhou, ``Tangent convolutions for
  dense prediction in 3d,'' in \emph{CVPR}, 2018, pp. 3887--3896.

\bibitem{kochanov2020kprnet}
D.~Kochanov, F.~K. Nejadasl, and O.~Booij, ``Kprnet: Improving projection-based
  lidar semantic segmentation,'' \emph{arXiv preprint arXiv:2007.12668}, 2020.

\bibitem{gerdzhev2020tornado}
M.~Gerdzhev, R.~Razani, E.~Taghavi, and B.~Liu, ``Tornado-net: multiview total
  variation semantic segmentation with diamond inception module,'' \emph{arXiv
  preprint arXiv:2008.10544}, 2020.

\bibitem{semantickitti}
J.~Behley, M.~Garbade, A.~Milioto, J.~Quenzel, S.~Behnke, C.~Stachniss, and
  J.~Gall, ``{SemanticKITTI: A Dataset for Semantic Scene Understanding of
  LiDAR Sequences},'' in \emph{ICCV}, 2019.

\bibitem{shi2020spsequencenet}
H.~Shi, G.~Lin, H.~Wang, T.-Y. Hung, and Z.~Wang, ``Spsequencenet: Semantic
  segmentation network on 4d point clouds,'' in \emph{CVPR}, 2020, pp.
  4574--4583.

\bibitem{kirillov2019panoptic}
A.~Kirillov, K.~He, R.~Girshick, C.~Rother, and P.~Doll{\'a}r, ``Panoptic
  segmentation,'' in \emph{CVPR}, 2019, pp. 9404--9413.

\bibitem{behley2020benchmark}
J.~Behley, A.~Milioto, and C.~Stachniss, ``A benchmark for lidar-based panoptic
  segmentation based on kitti,'' \emph{arXiv preprint arXiv:2003.02371}, 2020.

\bibitem{milioto2020iros}
A.~Milioto, J.~Behley, C.~McCool, and C.~Stachniss, ``{LiDAR Panoptic
  Segmentation for Autonomous Driving},'' 2020.

\bibitem{jiang2020pointgroup}
L.~Jiang, H.~Zhao, S.~Shi, S.~Liu, C.-W. Fu, and J.~Jia, ``Pointgroup: Dual-set
  point grouping for 3d instance segmentation,'' in \emph{CVPR}, 2020, pp.
  4867--4876.

\bibitem{shi2020pv}
S.~Shi, C.~Guo, L.~Jiang, Z.~Wang, J.~Shi, X.~Wang, and H.~Li, ``Pv-rcnn:
  Point-voxel feature set abstraction for 3d object detection,'' in
  \emph{CVPR}, 2020, pp. 10\,529--10\,538.

\bibitem{wang2020reconfigurable}
T.~Wang, X.~Zhu, and D.~Lin, ``Reconfigurable voxels: A new representation for
  lidar-based point clouds,'' \emph{Conference on Robot Learning}, 2020.

\bibitem{choy20194d}
C.~Choy, J.~Gwak, and S.~Savarese, ``4d spatio-temporal convnets: Minkowski
  convolutional neural networks,'' in \emph{CVPR}, 2019, pp. 3075--3084.

\end{thebibliography}

%
%
%
%




\end{document}